\documentclass[lettersize,journal]{IEEEtran}
\usepackage{amsmath,amsfonts}
\usepackage{algorithmic}
\usepackage{algorithm}
\usepackage{array}
\usepackage[caption=false,font=normalsize,labelfont=sf,textfont=sf]{subfig}
\usepackage{textcomp}
\usepackage{stfloats}
\usepackage{url}
\usepackage{verbatim}
\usepackage{graphicx}
\usepackage{cite}
\usepackage{booktabs}
\usepackage{multirow}
\usepackage{multicol}
\usepackage{microtype}
\usepackage{arydshln}
\usepackage[table]{xcolor}
\usepackage{csquotes}
\usepackage{colortbl}

\begin{document}

\title{Beyond a Single Query: Multi-Perspective Auxiliary Representation Learning for Moment Retrieval}

\author{Mengnan~Liu,
        Sanping Zhou,~\IEEEmembership{Member,~IEEE,} 
        Kun Xia,
        Jiaxuan Lin,
        % Nanning~Zheng,~\IEEEmembership{Fellow,~IEEE,}
        Gang~Hua,~\IEEEmembership{Fellow,~IEEE,}
        and~Le~Wang,~\IEEEmembership{Senior Member,~IEEE}
       
\thanks{M. Liu, S. Zhou, J. Lin and L. Wang are with the State Key Laboratory of Human-Machine Hybrid Augmented Intelligence, Institute of Artificial Intelligence and Robotics, Xi'an Jiaotong University, Xi'an, Shaanxi 710049, China. E-mail: Liumn@stu.xjtu.edu.cn, \{spzhou,lewang\}@mail.xjtu.edu.cn. (Corresponding author: Sanping Zhou.)

\IEEEcompsocthanksitem K. Xia is with the School of Computer Science and Technology, Xi'an Jiaotong University, Xi'an, Shaanxi 710049, China. E-mail: kunxia@xjtu.edu.cn
% \IEEEcompsocthanksitem Y. Dong is with the School of Software Engineering, Xi’an Jiaotong University, Xi’an 710049, China. E-mail: yhdong36@xjtu.edu.cn
\IEEEcompsocthanksitem G. Hua is with the Amazon Alexa AI. E-mail: ganghua@gmail.com.
}}


\maketitle

\begin{abstract}
Moment Retrieval (MR) aims to localize the temporal segment in a video that corresponds to a given textual description. However, a textual query often provides only a sparse semantic description of the target event, whereas the video contains rich and fine-grained visual variations. This mismatch creates an inherent gap between textual semantics and visual content, making a single textual input insufficient for precise cross-modal alignment. As a result, when multiple temporally adjacent or semantically similar segments exist, the model may struggle to distinguish the target moment from confusing candidates.
To address this challenge, we propose a Multi-Perspective Auxiliary Representation (MPAR) framework, which aims to bridge this alignment gap by enriching a single textual input with complementary visual cues from multiple perspectives.
Specifically, MPAR consists of three key components: (1) Auxiliary Representation Initialization (ARI) constructs multiple learnable representation slots from the original query, preserving the query intent while preparing flexible carriers for subsequent visual cue injection; (2) Visual-Induced Semantic Structuring (VISS) injects target-specific visual attributes into each representation while preserving semantic coherence with the original query; and (3) Multi-Perspective Semantic Aligner (MPSA) aligns each auxiliary representation with corresponding video regions to enhance discriminative localization. 
By constructing multi-perspective auxiliary representations and guiding the model to attend to relevant visual regions across these representations, our model achieves more accurate moment localization. Extensive experiments on MR benchmarks demonstrate that our MPAR  outperforms state-of-the-art methods, validating the effectiveness of MPAR.
\end{abstract}


\section{Introduction}

\IEEEPARstart{M}{oment} Retrieval (MR) aims to localize a temporal segment in an untrimmed video given a text query. Owing to its practical value, MR serves as a cornerstone technique for a variety of applications, including video summarization~\cite{park2025language, islam2025towards}, anomaly detection~\cite{yan2024anomalysd, enayati2024time, zhou2025optimization}, and video editing~\cite{he2024llms, sandoval2025editduet}. 
Unlike conventional moment localization~\cite{zeng2019graph, yang2020revisiting, liu2024stepwise} or video segmentation tasks~\cite{schmidt2025prior2former, zhou2025hypidecoder}, MR enables flexible natural language queries to describe target moments from diverse semantic perspectives, leading to inherently rich but highly ambiguous cross-modal associations.
Therefore, the core challenge of MR lies in learning fine-grained and context-aware alignment between textual semantics and dynamic visual content, so as to accurately distinguish target moments from visually and temporally similar candidates in complex videos.

\begin{figure}[tb]
  \centering
  \includegraphics[width=\linewidth]{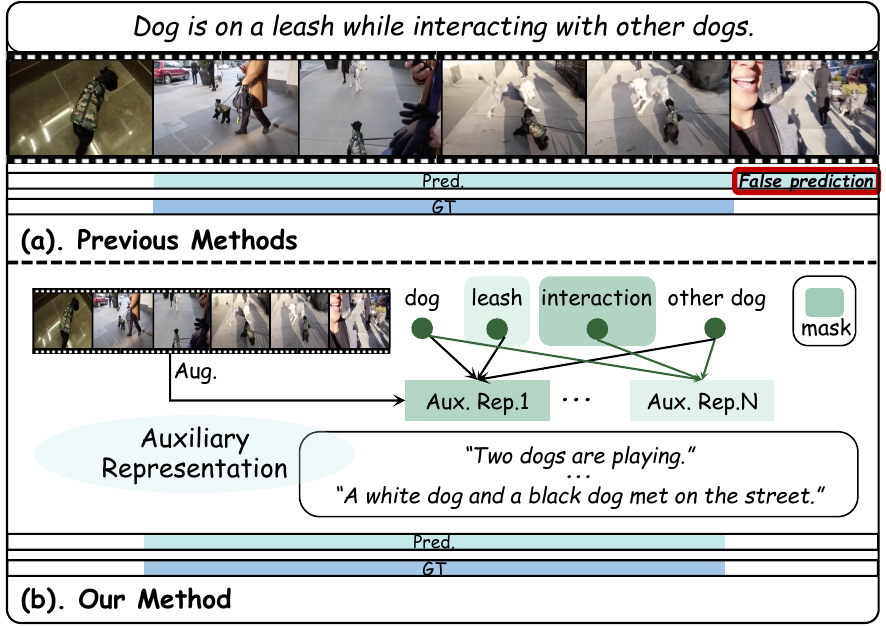}
  \caption{Illustration of the motivation behind our approach. A single textual query often provides only sparse semantic cues, creating a gap between the textual description and the rich visual content in complex videos. MPAR constructs multi-perspective auxiliary representations that preserve the query intent and incorporate target-related visual cues, thereby providing richer semantic guidance to alleviate this text-video gap.}
  \label{fig:intro}
  % \vspace{-0.5cm}
\end{figure}

Existing MR methods~\cite{xu2019multilevel, zhang2020learning, sun2024tr, zhou2025devil} typically rely on a single-sentence query to guide video grounding. 
While a textual query typically describes the main subject and action, accurate localization often relies on more comprehensive visual information that is implied by the event but not explicitly mentioned in the description, such as the specific scene or interacting objects.
Fig.~\ref{fig:intro} provides a concrete example. The query \enquote{a dog on a leash interacting with other dogs} offers one perspective, yet the actual event encompasses richer details like the number and colors of the dogs. Existing methods that depend on this single, sparse query are prone to overfitting to its specific wording (\textit{e.g.}, \enquote{leash}), while ignoring other discriminative visual attributes. 
This reliance on sparse textual semantics leads to brittle cross-modal alignment, causing failures when the emphasized cue is absent or shared across multiple candidate segments.

Building upon the above insight, we propose a novel framework named Multi-Perspective Auxiliary Representation (MPAR), which enriches a single textual query by generating multiple complementary representations. Each representation preserves the original query’s semantic intent while incorporating distinct visual cues that are implicitly present in the video but overlooked in the original description. These auxiliary representations serve as enriched semantic proxies, enabling the model to attend to diverse yet relevant visual regions across the video. By explicitly bridging textual semantics with visual context through multi-perspective representations, our approach mitigates the risk of over-reliance on sparse textual cues and improves the robustness of cross-modal alignment for moment localization.

Specifically, our MPAR framework consists of three core components: (1) Auxiliary Representation Initialization (ARI), (2) Visual-Induced Semantic Structuring (VISS), and (3) Multi-Perspective Semantic Aligner (MPSA).
% In the initialization stage, we generate multiple auxiliary text representations from the original query through token dropout and length-aware transformation. This process introduces diversity in semantic granularity and expression, enabling the model to view the same query from different linguistic perspectives.
% To incorporate visual context while maintaining semantic consistency, we propose the Visual-Induced Semantic Structuring (VISS) module. VISS operates in two complementary steps: it first extracts global contextual information from the video and injects it into each auxiliary representation to preserve global semantic coherence. It then selectively incorporates localized visual cues by measuring similarity between textual and visual features, allowing each representation to focus on different visually relevant regions.
% Furthermore, we design a Multi-Perspective Semantic Aligner (MPSA) to compute alignment between auxiliary representations and video segments from multiple semantic views. By aggregating alignment signals across these perspectives, MPSA enhances the discriminability of temporal localization and improves robustness against ambiguous visual patterns, effectively narrowing the gap between textual semantics and video content.
% Specifically, our MPAR framework comprises three core components: (1) Auxiliary Representation Initialization, (2) Visual-Induced Semantic Structuri, and (3) Multi-Perspective Semantic Aligner.
In the initialization stage, we construct multiple learnable auxiliary representation slots from the original query. These slots preserve the original query intent and provide flexible carriers for subsequent visual-induced semantic enrichment.
To inject target-aware visual cues while preserving semantic coherence, we propose the Visual-Induced Semantic Structuring (VISS) module. VISS consists of two complementary units:
The Visual Semantic Alignment Unit extracts global thematic information from the video and integrates it into each auxiliary representation, ensuring all variants remain anchored to the same underlying subject semantics.
The Semantic Diversification Generator enriches each representation with localized visual attributes. It identifies target-relevant video segments by measuring similarity between text embeddings and visual features, then incorporates these selected visual tokens into auxiliary representations via cross-modal attention. This enables each representation to emphasize distinct visual aspects of the target, broadening semantic coverage without losing relevance.

Furthermore, we design a Multi-Perspective Semantic Aligner (MPSA) to compute fine-grained alignment scores between auxiliary representations and video segments. By aggregating multi-perspective semantic evidence, MPSA produces semantic-aware visual feature maps that highlight regions most relevant to the query and its auxiliary variants. This effectively bridges the gap between sparse language descriptions and rich video content, enabling more discriminative moment localization.

To summarize, our contributions are as follows:
\begin{itemize}
    \item We develop a novel moment retrieval framework that generates multi-perspective auxiliary representations from a single query, leveraging visual cues to mitigate semantic ambiguity and enhance localization robustness.
    \item We propose a Visual-Induced Semantic Structuring module that enriches auxiliary representations with target-specific visual attributes while maintaining semantic coherence with the original query.
    \item We introduce a Multi-Perspective Semantic Aligner module that computes fine-grained cross-modal correlations, enabling precise alignment between the enriched representations and relevant video segments.
    \item Extensive experiments on multiple moment retrieval benchmarks demonstrate the effectiveness and generalization ability of our framework, while comprehensive ablation studies and analyses validate the contributions of each proposed component.
\end{itemize}

\section{Related Work}

% \subsection{Moment Retrieval}
% Moment Retrieval (MR) refers to the task of localizing the video segment in an untrimmed video that semantically corresponds to a provided language description. Since being first introduced in \cite{gao2017tall}, MR has emerged as a key problem in video-language understanding, where existing methods generally fall into two major paradigms.
% Proposal-based methods \cite{gao2017tall, xu2019multilevel, zhang2020learning} first produce candidate segments via sliding windows, proposal networks, or 2D temporal maps, and then rank them based on their similarity to the query. While effective, they usually involve high computational overhead and are sensitive to proposal quality.
% To overcome these limitations, proposal-free methods \cite{ghosh2019excl, zhang2021natural, mun2020local} predict start and end boundaries directly, eliminating the need for explicit proposals and simplifying inference. Building on DETR \cite{carion2020end}, recent Transformer-based approaches reformulate MR as a set prediction problem, enabling end-to-end localization without handcrafted proposals or post-processing \cite{moon2023query, sun2024tr, lei2021detecting, yang2024task, zhou2025devil}. Although efficient, these models often generate overlapping or redundant predictions due to the absence of task-specific constraints. To mitigate this, techniques such as boundary refinement and attention regularization have been proposed to guide the model and improve temporal precision.

\subsection{Moment Retrieval}
Moment Retrieval (MR) aims to localize a temporal segment in an untrimmed video that semantically corresponds to a given language description. Since its introduction in~\cite{gao2017tall}, MR has become an important task in video-language understanding. Existing methods can be broadly divided into proposal-based and proposal-free paradigms. Proposal-based methods~\cite{gao2017tall, xu2019multilevel, zhang2020learning, lin2020moment} first generate candidate temporal segments through sliding windows, proposal networks, or dense temporal maps, and then rank them according to their cross-modal matching scores with the query. Although effective, these methods are usually computationally expensive and highly dependent on the quality of generated proposals. In contrast, proposal-free methods~\cite{ghosh2019excl, zhang2021natural, mun2020local, sun2021maban, yang2021local} directly predict the start and end boundaries of the target moment, which simplifies the localization pipeline and improves inference efficiency. In addition to conventional single-video grounding, Paul et al.~\cite{paul2021text} further study text-based localization of moments in a video corpus, extending MR to a more practical large-scale retrieval scenario.

Recently, Transformer-based approaches have reformulated MR as an end-to-end set prediction problem inspired by DETR~\cite{carion2020end}, avoiding handcrafted proposals and complex post-processing~\cite{moon2023query, sun2024tr, lei2021detecting, yang2024task, zhou2025devil}. These methods benefit from global temporal modeling and flexible query-based decoding, but they may still generate redundant predictions or struggle with fine-grained semantic alignment when visually similar moments appear in the same video. To further improve video-language interaction, recent studies also explore cross-modal neural architecture search~\cite{yang2022video} and caption-assisted multimodal large language models~\cite{xie2025caption}, showing the importance of richer semantic modeling for MR. Despite these advances, most existing methods still rely on a single textual representation to guide localization. Such a representation may be insufficient to capture diverse semantic focuses and fine-grained visual attributes in complex videos. In contrast, our method constructs multi-perspective auxiliary representations from the original query and enriches them with target-specific visual cues, enabling more robust and discriminative cross-modal alignment for accurate moment localization.

\begin{figure*}[tb]
  \centering
  \includegraphics[width=\textwidth]{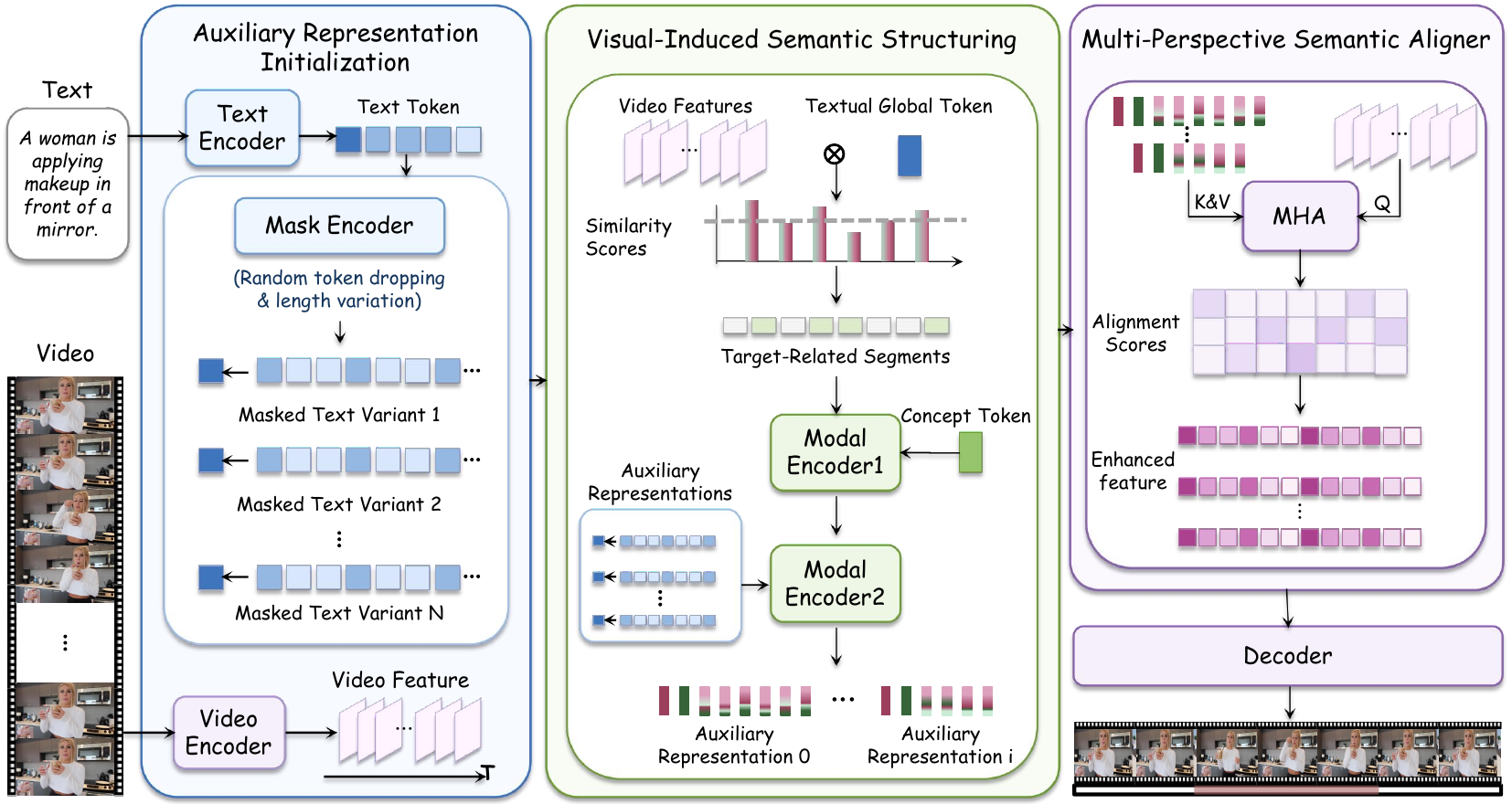}
\caption{Overview of the proposed MPAR framework. Given an input video and its corresponding text query, MPAR first constructs a set of auxiliary representations from the original description. Rather than introducing independent semantics, these representations serve as learnable placeholders that preserve the query intent and provide flexible slots for subsequent visual cue injection. Then, the generated representations are combined with global semantic tokens and fed into the Visual-Induced Semantic Structuring module. In this module, they interact with video features to preserve a consistent target subject while incorporating target-related visual cues, thereby enriching the textual semantics with visually grounded attributes.  The enriched representations are further aligned with temporal video features by MPSA, narrowing the gap between textual semantics and visual observations for accurate moment localization.}

  \label{fid1}
  % \vspace{-1cm}
\end{figure*}

\subsection{Representation Refinement}
Representation refinement has been widely studied to improve the robustness and generalization ability of language and vision-language models~\cite{lu2019vilbert}. Early refinement methods mainly operate at the input level, where diverse textual variants are generated through token replacement, deletion, insertion, or back-translation~\cite{wei2019eda, edunov2018understanding}. These methods can increase surface-level diversity while largely preserving the original semantics, but they are often limited to discrete word-level transformations and may introduce noisy or task-irrelevant expressions. To overcome this limitation, recent latent-space refinement methods construct semantic variations directly in the hidden representation space. For example, SimCSE~\cite{gao2021simcse} introduces stochastic perturbations through dropout to learn robust sentence embeddings, while E-Mix~\cite{jindal2020augmenting} performs representation interpolation to improve model generalization. These methods demonstrate that hidden-space refinement can provide smoother and more flexible semantic variations than explicit textual augmentation.

Inspired by this work, our method performs representation refinement in the hidden space to construct multiple auxiliary textual representations from the original query. However, unlike existing refinement methods that mainly treat generated variants as homogeneous augmented samples, our goal is to produce multi-perspective representations for cross-modal grounding. Specifically, we encourage all auxiliary representations to preserve thematic consistency with the original query, ensuring that they refer to the same underlying target event. Meanwhile, controlled variation is introduced to make each representation emphasize different semantic aspects and visually relevant cues. This design allows the model to avoid over-reliance on a single textual expression and provides complementary semantic views for subsequent video-language alignment. By balancing semantic consistency and controllable diversity, our refinement strategy enhances discriminative cross-modal alignment for moment retrieval.

% \subsection{Representation Refinement}
% Representation refinement has been extensively explored to improve the robustness and generalization of language models~\cite{lu2019vilbert}. Traditional refinement methods~\cite{wei2019eda, edunov2018understanding}, including token replacement~\cite{wei2019eda} and back-translation~\cite{edunov2018understanding}, typically generate diverse surface-level variations while retaining semantic consistency. More recently, latent-space refinement methods, such as SimCSE~\cite{gao2021simcse} and E-Mix~\cite{jindal2020augmenting}, construct semantic variations by applying stochastic perturbations or representation interpolation within the hidden space.
% In this vein, our approach performs representation refinement in the hidden space to produce multiple textual representations. However, unlike existing methods that treat these representations as homogeneous variants, we explicitly enforce all variants to share a thematic consistency, while allowing controllable variations. This design ensures that all generated auxiliary representations refer to the same target, while each captures unique and visually grounded semantic cues, thereby enabling more discriminative and consistent cross-modal alignment in subsequent visual reasoning tasks.

\section{Methodology}
\subsection{Overview}
In this section, we introduce the Multi-Perspective Auxiliary Representation (MPAR) framework. The core idea of MPAR is to construct multiple query-preserving auxiliary representation slots and enrich them with target-specific visual cues, thereby compensating for visual details that are not explicitly expressed in the original text. Each auxiliary representation is enriched with additional visual cues, enabling the model to interpret the same target from multiple perspectives.
% , which in turn improves the accuracy and robustness of moment localization.

As illustrated in Fig.~\ref{fid1}, our framework consists of three key components: 1) the Auxiliary Representation Initialization~(ARI) (Sec.\ref{Auxiliary Representation Initialization}), 2) the  Visual-Induced Semantic Structuring~(VISS) (Sec.\ref{Visual-Induced Semantic Structuring}), where we detail the Visual Semantic Alignment Unit and Semantic Diversification Generator modules, and 3) the Multi-Perspective Semantic Aligner (MPSA) (Sec.\ref{Multi-Perspective Semantic Aligner}).

\subsection{Multimodal Feature Extraction}
Given an untrimmed video $V=\{v_t\}_{t=1}^{L_{v}}$ with $L_{v}$ frames and a corresponding natural language description $T=\{t_n\}_{n=1}^{L_{t}}$ containing $L_{t}$ words, moment retrieval~(MR) aims to precisely predict the temporal span $\mathbf{m} = (m_c, m_{\sigma})$ that best aligns with the given description, where $m_c$ and $m_{\sigma}$ denote the center time and duration of the target moment, respectively.
\newline
\textbf{Visual Features.} Following the literature~\cite{lei2021detecting}, the video is first divided into non-overlapping clips according to a certain time interval. Then the pre-trained video encoder is utilized to extract clip-level visual features $\mathbf{F}_{v} = [f_{v}^{1}, f_{v}^{2},..., f_{v}^{N_v}] \in \mathbb{R}^{N_v \times d_{v}}$, where $N_v$ and $d_v$ are the number of clips and the visual feature dimension, respectively. 
\newline
\textbf{Textual Features.} For a natural language query, we use the pre-trained textual encoder to extract textual features $\mathbf{F}_t = [f_{t}^{1}, f_{t}^{2},..., f_{t}^{N_t}] \in \mathbb{R}^{N_t \times d_t}$, where $N_t$ and $d_{t}$ indicate the number of tokens and the textual feature dimension, respectively.

\subsection{Auxiliary Representation Initialization}
\label{Auxiliary Representation Initialization}

We first construct multi-perspective auxiliary representations from the original textual description. Instead of directly introducing new semantic meanings, ARI aims to build a set of learnable semantic placeholders that preserve the original query intent and provide representation slots for subsequent visual cue injection. 
Specifically, a subset of semantic tokens is randomly dropped and the remaining tokens are adaptively transformed into auxiliary representations with different token lengths, providing flexible representation spaces for subsequent visual grounding.
This process does not independently alter the target semantics, but creates multiple flexible representation spaces that can later absorb complementary target-specific visual cues. In this way, each auxiliary representation remains semantically anchored to the original query while providing a distinct carrier for subsequent visual grounding.

% We first construct multi-perspective auxiliary representations from the original textual description. To encourage semantic diversity, the textual features are fed into a mask encoder to generate multiple Auxiliary Representations. Specifically, a subset of semantic tokens is randomly dropped and the remaining tokens are adaptively re-ordered in length, producing multiple auxiliary representations with different granularities and focuses.
% In this way, each auxiliary representation preserves the original semantic intent while emphasizing a distinct subset of discriminative cues, thereby providing complementary perspectives for subsequent visual grounding.

% , laying the foundation for subsequent visual guidance and semantic alignment.
% We begin by initializing the auxiliary representations from the textual features. To encourage semantic diversity, we randomly drops partial semantic tokens and performs a length transformer to generate multiple textual variants with distinct semantic focuses.

Formally, given the tokens of the input text feature $\mathbf{F}_t = [f_{t}^{1}, f_{t}^{2},..., f_{t}^{N_t}] \in \mathbb{R}^{N_t \times d_t},$
we construct $N_a$ auxiliary representation sets $\{\mathbf{A}^i\}_{i=1}^{N_a}$, each with a pre-defined token length $k^i$.  
The $i$-th auxiliary representations are generated as,
\begin{equation}
\mathbf{A}^i = [\mathbf{a}_{1}^i, \ldots, \mathbf{a}_{k^i}^i] = \phi^i(\mathbf{D}^i \odot \mathbf{F}_t), \quad
\mathbf{A}^i \in \mathbb{R}^{k^i \times d_t},
\label{eq:attr_init}
\end{equation}
where $\phi^i \in \mathbb{R}^{k^i \times N_t}$ denotes the \emph{mask encoder}, $\mathbf{D}^i \in \{0,1\}^{N_t \times d_t}$ is a dropout mask with a dropping probability $p^i$, and $\odot$ represents the Hadamard product.  
This step creates multiple query-preserving token configurations, which provide flexible representation slots for subsequent visual-induced semantic enrichment.

% To maintain consistency of the target subject across all generated expressions, we identify a \textbf{subject token} $\mathbf{s}$ from $\bar{T}_0$ via a linear projection followed by a straight-through Gumbel-Softmax~\cite{jang2016categorical} operation:
% \begin{equation}
% \mathbf{s} = \text{GumbelSoftmax}(\mathbf{W}_s \bar{T}_0),
% \label{eq:subject_token}
% \end{equation}
% where $\mathbf{W}_s$ is a learnable projection matrix.  
% The subject token $\mathbf{s}$ is shared across all auxiliary representations to provide a unified semantic reference.
Meanwhile, in auxiliary representation modeling, we aim to construct multiple query-preserving yet visually extensible auxiliary representations, enabling the model to incorporate complementary visual cues while maintaining subject consistency.
However, if these auxiliary representations are generated independently, they may lack global consistency, causing the expressions to deviate from the original semantic subject or even suffer from semantic drift.
To maintain diversity while ensuring a unified semantic among all auxiliary representations, we introduce a global anchor $\mathbf{z}_{[\text{global}]}^i$. These anchors enable multiple auxiliary representations to capture their respective semantic attributes while maintaining semantic consistency without diverging from each other.
% These anchors enable the model to capture diverse semantic attributes through multiple auxiliary representations while effectively preventing drift from the original semantics.

Finally, we prepend a global anchor  $\mathbf{z}_{[\text{global}]}^i$ to form the $i$-th initial auxiliary representation as follows,
\begin{equation}
\mathbf{Z}^i = [\mathbf{z}_{[\text{global}]}^i, \mathbf{A}^i] \in \mathbb{R}^{(k^i + 1) \times d_t},
\label{eq:latent_init}
\end{equation}
where the global anchor $\mathbf{z}_{[\text{global}]}^i$ is designed as a learnable parameter, which is continuously updated through cross-modal alignment objectives. This allows $\mathbf{z}_{[\text{global}]}^i$ to distill the semantic elements shared across auxiliary representations and serve as a global semantic reference.
Each initial $\mathbf{Z}^i$ thus represents a distinct auxiliary representation containing semantic variations but sharing a consistent subject reference from the original description.
\subsection{Visual-Induced Semantic Structuring}
\label{Visual-Induced Semantic Structuring}
In this section, we introduce two key components: (1) Visual-Semantic Alignment Unit and (2) Semantic Diversification Generator. These modules are designed to jointly maintain a consistent high-level visual subject across auxiliary representations while enabling each representation to focus on complementary target-specific visual attributes.
Specifically, the alignment unit preserves global semantic coherence between text and visual content, ensuring all representations remain grounded on the same underlying event. In contrast, the diversification generator injects localized and discriminative visual cues, encouraging each auxiliary representation to capture distinct aspects of the target semantics.

Here, $z_{global}$ maintains query-level consistency, $z_v$ provides a shared visual subject anchor, $z_t$ guides target-related segment selection, and concept tokens $\mathbf{C}$ serve as visual cue carriers for local semantic enrichment.
% In this section, we introduce two key components: 1) Visual-Semantic Alignment Unit and 2) Semantic Diversification Generator. These components aim to ensure that different auxiliary representations maintain a consistent visual subject while incorporating target-specific visual attributes.

% \label{latent query}
% The subject distributor ensures that the target subject is consistently referenced. Our method is inspired by the example in Fig. 1a, where the target can be described with diverse expressions, all referring to the same subject, "man," but highlighting different attributes ("left", "denim," and "older"). We believe that by collectively considering these attributes, the target can be identified more precisely. However, since the input text provides only sparse attributes, we extract additional attributes from the visual domain to refine the representation further.
% The subject tokens $\{\mathbf{s}^{i}\}_{i=1}^{N}$ across all latent expressions $\{\mathbf{Z}^{i}\}_{i=1}^{N}$ may no longer represent the same subject or may have arbitrary semantics. To address this, we realign these tokens by exchanging the subject tokens $\{\mathbf{s}^i\}_{i=1}^{N}$ in $\{\mathbf{Z}\}_{i=1}^{N}$ into the visual subject token $\mathbf{s}^{V}$ in $\mathbf{v}$,
% \begin{equation}
% \label{latent init}
% \mathbf{Z}{i} \leftarrow [\mathbf{z}_{[cls]}^{i}, \mathbf{s}^{V}, \mathbf{A}^{i}] \in \mathbb{R}^{(k^i+2)\times d}.
% \end{equation}
% This ensures that all $\{\mathbf{Z}{i}\}_{i=1}^{N}$ consistently share the same visual subject throughout the subsequent layers.
\subsubsection{Visual-Semantic Alignment Unit}

To ensure semantic coherence across auxiliary representations, we introduce the Visual Semantic Alignment Unit.
This design is motivated by the observation that a single visual event can be described using diverse linguistic expressions that emphasize different aspects of the same underlying scene (e.g.,  \enquote{Dog is on a leash while interacting with other dogs.},  \enquote{Two dogs are playing.} or \enquote{A white dog and a black dog met on the street.}), which emphasize distinct visual attributes but share a common visual subject. Although these descriptions differ in surface form, they correspond to a shared latent semantic anchor in the visual content.
Based on this insight, the alignment unit encourages all auxiliary representations to remain grounded on a consistent visual subject, enabling the model to reduce semantic drift while improving robustness to linguistic variability in moment localization.

% To ensure that all auxiliary representations consistently preserve semantic coherence with the original query, we introduce the Visual Semantic Alignment Unit.  
% This module is motivated by the observation that a target segment can be described through multiple linguistic expressions (e.g.,  \enquote{Dog is on a leash while interacting with other dogs.},  \enquote{Two dogs are playing.} or \enquote{A white dog and a black dog met on the street.}), which emphasize distinct visual attributes but share an identical subject. By aligning these expressions to a common visual subject, we enable the model to perceive the target more precisely and robustly.

% However, since each auxiliary representation $\mathbf{Z}^i$ is initialized independently, its subject may drift toward inconsistent or even arbitrary semantics. To rectify this misalignment, 
To achieve this goal, VSAU introduces a unified visual subject token $\mathbf{z}_{v}$ extracted from the video domain, which serves as a global visual anchor shared across all representations and is formally defined as follows,
\begin{equation}
\mathbf{z}_{v} = \boldsymbol{\alpha} \mathbf{F}_{v}, \quad \boldsymbol{\alpha} = \text{softmax}\Big(\frac{\mathbf{q}_v \mathbf{F}_{v}^\top}{\sqrt{d_v}}\Big),
\label{eq:sV}
\end{equation}

where $\mathbf{q}_v$ is a learnable parameter that serves as a query vector to extract the global representation of visual features.
% , while $d$ is a scaling factor used to normalize the attention scores and prevent numerical instability.

Formally, the auxiliary representation is updated as,
\begin{equation}
\mathbf{Z}^i = [\mathbf{z}_{[\text{global}]}^{i}, \mathbf{z}_{v}, \mathbf{A}^{i}], \quad
\mathbf{Z}^i \in \mathbb{R}^{(k^i + 2) \times d_t}.
\label{eq:vsau}
\end{equation}

Subsequently, all auxiliary representations $\{\mathbf{Z}^i\}_{i=1}^{N_a}$ share a consistent visual subject $\mathbf{z}_{v}$, while retaining diverse attribute tokens $\mathbf{A}^i$ that attend to distinct semantic cues.
This design effectively grounds textual variations in a shared visual context, bridging cross-modal semantics and ensuring subject coherence throughout the subsequent reasoning process.

\subsubsection{Semantic Diversification Generator}
Although ARI constructs multiple learnable representation slots, these slots are initialized from the same textual description and thus may still lack explicit fine-grained visual distinctions. Since the target moment often contains localized visual attributes that are implicitly reflected in the query, it is necessary to selectively introduce target-related visual cues from the video. To this end, we design the Semantic Diversification Generator (SDG), which mines discriminative visual concepts and injects them into auxiliary representations, enabling different representations to absorb complementary visual attributes while preserving the original query semantics.

To capture fine-grained and diverse visual cues within auxiliary representations, we introduce a set of target-specific visual tokens into the representation space $\mathbf{A}^{i} = [\mathbf{a}_{1}^{i}, \dots, \mathbf{a}_{k^{i}}^{i}]$.
Specifically, we initialize $N_c$ concept tokens $\mathbf{C} \in \mathbb{R}^{N_c \times d_v}$ using orthogonal initialization, encouraging each token to encode distinct visual semantics and preventing representation collapse. 
To provide global contextual guidance, we further derive a text-aware global token $\mathbf{z}_t$ from the textual feature sequence $\mathbf{F}_{t}$, which serves as a semantic anchor for subsequent visual-token interaction and alignment.

% To capture distinct details in the generated auxiliary representations, we incorporate target-specific visual cues into representation tokens $\mathbf{A}^{i} = [\mathbf{a}_{1}^{i}, \dots, \mathbf{a}_{k^{i}}^{i}]$. We begin by initializing $N_c$ concept tokens $\mathbf{C} \in \mathbb{R}^{N_{c} \times d_v}$ using orthogonal initialization to ensure their distinctiveness. 
% Next, we derive the text global token $\mathbf{z}_t$ from the textual features $\mathbf{F}_{t}$ as follows,
\begin{equation}
\label{eq:tc}
\mathbf{z}_t = \boldsymbol{\gamma} \mathbf{F}_{t}, \quad \boldsymbol{\gamma} = \text{softmax}\Big(\frac{\mathbf{q}_t \mathbf{F}_{t}^\top}{\sqrt{d_t}}\Big),
\end{equation}
where $\mathbf{q}_t$ is a learnable parameter that serves as a query vector to extract the global representation of textual features.

We then choose target-related segments $\mathbf{V}_{\text{t}}$ from the visual features $\mathbf{F}_{v} = [f_{v}^{1}, f_{v}^{2},..., f_{v}^{N_v}]$ by thresholding their similarities with the textual global token $\mathbf{z}_t$ as follows, 
\begin{equation}
\label{eq:vtr}
\mathbf{V}_{t} = \{\, f_{v}^{i} \mid f_{v}^{i}\mathbf{z}^\top _{t} \ge \theta,\; i \in \{1, \ldots, N_v\} \,\} \in \mathbb{R}^{N_{\text{r}} \times d_v},
\end{equation}
where $\theta$ denotes the similarity threshold, and $N_{\text{r}}$ is the number of selected target-related segments. 
% We set $\theta$ to the mean similarity value so that most visual segments are retained, which reduces the dependence on textual input while preserving contextual information beyond the target region.
Based on the selected target-related segments $\mathbf{V}_{\text{t}}$, we retrieve visual concept cues by performing a Modal Encoder (i.e., attention) over their corresponding concept tokens $\mathbf{C}$, yielding the visual concept token $\mathbf{C}_{v}$,
\begin{equation}
\label{eq:cv_hadamard}
\mathbf{C}_{v} =  \boldsymbol{\beta} \mathbf{V}_{\text{t}},
\quad
\boldsymbol{\beta} = \text{Softmax}\!\left(\mathbf{C} \mathbf{V}_{\text{t}}^{\top}\right),
\end{equation}
where the softmax normalizes attention weights over the selected target-related temporal segments.
We then incorporate these unique visual concepts $\mathbf{C}_{v}$ into the concatenated auxiliary representation tokens $\mathcal{A} = [\mathbf{A}^1, \ldots, \mathbf{A}^N] \in \mathbb{R}^{N_a \times d_t}$ (where $N_a$ is the total number of auxiliary representations $\{\mathbf{Z}^i\}_{i=1}^{N_a}$) via another Modal Encoder (i.e., attention), where we normalize the attention score over the representation tokens,
% \begin{equation}
% \label{eq:ainject}
% \tilde{\mathbf{A}} = \tilde{W} \mathbf{C}^{V} \in \mathbb{R}^{N_a \times d},
% \end{equation}
% where
% \[
% \tilde{W}_{n,k} = \frac{e^{\tilde{M}_{n,k}}}{\sum_{n'=1}^{N_{a}} e^{\tilde{M}_{n',k}}}, 
% \quad
% \tilde{M} = \tilde{\mathbf{A}} (\mathbf{C}^{V})^\top.
% \]
\begin{equation}
\label{eq:ainject_refined}
\mathcal{A} = \mathbf{W} \mathbf{C}_{v},
\quad
\mathbf{W} = \text{Softmax}\!\left(\mathcal{A}\mathbf{C}_{v}^{\top}\right).
\end{equation}

By regularizing the attention distribution across representation tokens, we encourage competition in attending to visual concepts, which drives each token to specialize in a distinct subset of visual semantics.
These new concatenated attribute tokens $\mathbf{A}^i$ are then split back to comprise the auxiliary representation $\mathbf{Z}^i$,
\begin{equation}
\label{eq:zupdate}
\mathbf{Z}^{i} = [\mathbf{z}_{[\text{global}]}^{i},\, \mathbf{z}_{v},\, \mathbf{A}^{i}] \in \mathbb{R}^{(k^i + 2) \times d_t}.
\end{equation}
These updated auxiliary representations are then fed into the Multi-Perspective Semantic Aligner.

\subsection{Multi-Perspective Semantic Aligner}
\label{Multi-Perspective Semantic Aligner}

Previous methods~\cite{sun2024tr, moon2023correlation} typically rely solely on visual feature maps $\mathbf{F}_v$, which lack explicit semantic priors aligned with the textual query. As a result, the model is prone to focusing on visually salient but semantically irrelevant regions, leading to suboptimal temporal localization.
To address this limitation, we introduce the Multi-Perspective Semantic Aligner (MPSA), which leverages auxiliary representations as semantic priors to guide video feature modeling. Specifically, the auxiliary representations provide complementary cross-view semantic cues that enable more structured alignment between language and vision.

% As illustrated in Fig.~\ref{fid1}, MPSA adopts a multi-head cross-attention mechanism~\cite{vaswani2017attention}, where the visual feature map $\mathbf{F}_v \in \mathbb{R}^{N_v \times d_v}$ is used as the query, and the auxiliary representations $\mathbf{Z} \in \mathbb{R}^{N_a \times d_t}$ serve as keys and values. This design aggregates multi-perspective semantic information into each temporal location, producing a refined feature map $\mathbf{F}_s \in \mathbb{R}^{N_v \times d_v}$ that is explicitly enhanced with query-relevant semantic context for improved localization.

% In previous methods like~\cite{sun2024tr, moon2023correlation}, the visual feature map $\mathbf{F}_v$ lacks the semantic prior needed to identify the text-referred target, making localization easily distracted. 
% To address this issue, we incorporate auxiliary representations into the Multi-Perspective Semantic Aligner, which interact with visual features to establish cross-modal associations and supply multi-view semantic cues.
% These associations guide the model to focus on text-relevant visual details, reduce semantic drift, and improve temporal localization accuracy.

As illustrated in Fig.~\ref{fid1}, the MPSA module is based on a multi-head attention mechanism~\cite{vaswani2017attention}. Specifically, the visual feature $\mathbf{F}_v \in \mathbb{R}^{N_v \times d_v}$ serves as the \textit{query}, while the auxiliary representations $\mathbf{Z} \in \mathbb{R}^{N_a \times d_t}$ serve as both the \textit{keys} and \textit{values}. Through multi-head attention, semantic features from $\mathbf{Z}$ are aggregated for each temporal location in $\mathbf{F}_v$, resulting in a semantic feature map $\mathbf{F}_s \in \mathbb{R}^{N_v \times d_v}$ that is spatially aligned with $\mathbf{F}_v$.

We then project both feature maps into a shared semantic space using linear projections followed by $\ell_2$ normalization, obtaining $\mathbf{F}'$ and $\mathbf{F}_s'$. For each temporal location $t$, we compute a semantic alignment score to quantify the correspondence between visual and textual features,
\begin{equation}
\label{eq:semantic_score}
S(t) = \kappa \cdot \exp\Bigg(- \frac{\|\mathbf{F}'(t) - \mathbf{F}_s'(t)\|_2^4}{8\sigma^2} \Bigg).
\end{equation}
Here, $\kappa$ and $\sigma$ are learnable parameters, where $\kappa$ scales the score and $\sigma$ controls the decay of similarity. The alignment score $S(t)$ at each time step $t$ reflects the semantic relevance of the corresponding visual feature to the referring expression.

Finally, we modulate the visual feature map $\mathbf{F}_v$ with the alignment scores in a frame-wise manner to obtain the enhanced visual representation,
\begin{equation}
\label{eq:fv_modulate}
\hat{\mathbf{F}} = \mathbf{F}_v \cdot \mathbf{S},
\end{equation}
where $\hat{\mathbf{F}}$ represents the refined visual features emphasizing the regions semantically aligned with the textual query. Such modulation naturally suppresses irrelevant regions and enhances the representation of the referred segment, facilitating more accurate localization in subsequent stages.

\subsection{Training Objective}
We train the DETR head using a combination of cross-entropy (CE) loss, Generalized IoU (GIoU) loss~\cite{rezatofighi2019generalized}, and Smooth L1 loss, following the approach in TR-DETR~\cite{sun2024tr}.
% Specifically, the CE loss is applied to the IoU prediction branch to enhance localization quality estimation. 
The overall moment retrieval objective is formulated as,

\begin{equation}
\label{Query fusion}
\begin{split}
\mathcal{L}_{overall} =\ & \lambda_{L_{1}}\mathcal{L}_{1}(\mathbf{m}, \hat{\mathbf{m}}) 
+ \lambda_{IoU}\mathcal{L}_{IoU}(\mathbf{m}, \hat{\mathbf{m}}) \\
& + \lambda_{CE}\mathcal{L}_{CE}(y, \hat{y}) 
+ \lambda_{IoU}\mathcal{L}_{CE}(IoU, \hat{IoU}),
\end{split}
\end{equation}
where the ground-truth values include the moment center and duration $\mathbf{m}$, binary classification label $y$, and IoU score. The corresponding predictions are denoted by $\hat{\mathbf{m}}$, $\hat{y}$, and $\hat{\text{IoU}}$, respectively. $\lambda*$ are balancing parameters that are consistent with TR-DETR~\cite{sun2024tr}.

\begin{table*}[tb]
  \caption{Comparison of model performance on QVHighlights dataset.}
  \label{QVHighlights val}
  \centering
  \small
  \begin{tabular}{c c c c c c c c}

    \toprule
    \multirow{3}{*}{Method} & \multicolumn{5}{c}{MR} & \multicolumn{2}{c}{HD} \\  
    \cmidrule(lr){2-6}
    \cmidrule(lr){7-8}
    & \multicolumn{2}{c}{R1} & \multicolumn{3}{c}{mAP} & \multicolumn{2}{c}{$\geq$Very Good} \\ 
    \cmidrule(lr){2-3}
    \cmidrule(lr){4-6}
    \cmidrule(lr){7-8}
    & $@$0.5 & $@$0.7 & $@$0.5  & $@$0.75 & Avg. mAP & mAP & HIT$@$1 \\
    \midrule
    M-DET~\cite{lei2021detecting} (NeurIPS'2021) & 52.89 & 33.02 & 54.84 & 29.40 & 30.73 & 35.69 & 55.60 \\
    QD-DETR~\cite{moon2023query} (CVPR'2023) & 62.52 & 44.98 & 62.62 & 39.88 & 39.86 & 38.64 & 62.40 \\
    % UniVTG~\cite{lin2023univtg} (ICCV'2023) & 58.86 & 40.86 & 57.60 & 35.59 & 35.47 & 38.20 & 60.96 \\
    EaTR~\cite{jang2023knowing} (ICCV'2023) & 61.36 & 45.79 & 61.86 & 41.91 & 41.74 & 37.15 & 58.65 \\
    UniVTG~\cite{lin2023univtg} (ICCV'2023) & 66.65 & 52.19 & 64.37 & 46.68 & 45.18 & 40.18 & \underline{64.77}  \\
    TaskWeave~\cite{yang2024task} (CVPR'2024) & 64.26 & 50.06 & 65.39 & 46.47 & 45.38 & - & - \\
    % BAM-DETR~\cite{lee2024bam} (ECCV'2024) & 65.10 & 51.61 & 65.41 & 48.56 & \underline{47.61} & - & - \\
    CG-DETR~\cite{moon2023correlation} (arXiv'2024) & 67.35 & 52.06 & 65.57 & 45.73 & 44.93 & - & - \\  
    TR-DETR~\cite{sun2024tr} (AAAI'2024) & 67.10 & 51.48 & 66.27 & 45.09 & 46.42 & \underline{40.55} & \underline{64.77} \\
    LLMEPET~\cite{jiang2024prior} (ACM MM'2024) & 66.58 & 51.10 & - & - & 46.24 & - & - \\
        RGTR~\cite{sun2025diversifying} (AAAI'2025) &  \underline{67.68} & 52.90 &  \underline{67.38} &  \underline{48.00} & 46.95 & - & - \\
     TD-DETR~\cite{zhou2025devil} (ICCV'2025) &  65.88 &  \underline{53.67} & 66.43 & \textbf{49.86} & \textbf{49.05} & - & - \\
 
    \rowcolor{cyan!10}
    MPAR (Ours) & \textbf{69.61} & \textbf{53.68} & \textbf{68.15} & 47.30 & \underline{47.15} & \textbf{41.46} & \textbf{67.94} \\
    \bottomrule
  \end{tabular}
  % \vspace{-0.3cm}
\end{table*}

\begin{table*}[tb]
  \caption{Performance comparison on TACoS and Charades-STA.}
\label{TACoS and Charades-STA}
  \centering
  \resizebox{0.95\linewidth}{!}{
  \begin{tabular}{c c c c c c c c c}
  \toprule
   \multirow{2}{*}{Method} & \multicolumn{4}{c}{TACoS} & \multicolumn{4}{c}{Charades-STA} \\  
    \cmidrule(r){2-5}
    \cmidrule(r){6-9}
 & R$@$0.3 & R$@$0.5 & R$@$0.7 & mIoU &  R$@$0.3 & R$@$0.5 & R$@$0.7 & mIoU\\
\midrule
2D-TAN~\cite{tan2021augmented} (arXiv'2021)  & 40.0 & 28.0 & 12.9 & 27.2 & 58.8 & 46.0 & 27.5 & 41.3  \\
% VSLNet  & 35.5 & 23.5 & 13.2 & 25.0 & 60.3 & 42.7 & 24.1 & 41.6  \\
M-DETR~\cite{lei2021detecting} (NeurIPS'2021) & 38.0 & 24.7 & 12.0 & 25.5 & 65.8 & 52.1 & 30.6 & 45.5  \\
MABAN~\cite{sun2021maban} (TIP'2021) & - & - & - & - & -  & 56.3 & 32.3 & - \\
CMAS~\cite{yang2022video} (TIP'2022) & - &31.4 & 16.9 &28.0 & - & 48.4 & 29.4 & 43.6\\
MomentDiff~\cite{li2023momentdiff} (NeurIPS'2023) & 44.8 & 33.7 & - & - & - & 55.6 & 32.4 & -  \\
% UniVTG~\cite{lin2023univtg} () & 51.4 & 35.0 & 17.4 & 33.6 & 70.8 & 58.0 & 35.7 & 50.1  \\
% EaTR~\cite{jang2023knowing} & - & - & - & - & 68.5 & 44.9 & 35.7 & 50.1  \\
% UniVTG  & 51.4 & 35.0 & 17.4 & 33.6 & 70.8 & 58.0 & 35.7 & 50.1  \\
UniVTG~\cite{lin2023univtg} (ICCV'2023) & - & - & - & - & 68.0 & 57.2 & 36.1 & - \\
CG-DETR~\cite{moon2023correlation} (arXiv’2024)  & 52.2 & 39.6 & 22.2 & 36.5 & 70.4 & 58.4 & 36.3 & 50.1 \\
% TR-DETR & - & - & - & - & 70.7 & 56.2 & 32.5 &  48.8 \\
TR-DETR~\cite{sun2024tr} (AAAI'2024) & - & - & - & - & - & 57.6 & 33.5 & -  \\
LLMEPET~\cite{jiang2024prior} (ACM MM'2024) & 52.7 & - & 22.8 & 36.6 & 70.9 & - & 36.5 & \underline{50.3} \\
% BAM-DETR~\cite{lee2024bam} (ECCV'2024) & \textbf{56.7} & \textbf{41.5} & \textbf{26.7} & \textbf{39.3} & 72.9 & 59.9 & 39.3 & \underline{52.3} \\
RGTR~\cite{sun2025diversifying} (AAAI'2025) & \underline{53.0} & \underline{40.3} & \underline{24.3} & \underline{37.4} & \underline{72.0} & 57.9 & 35.2 & \underline{50.3} \\
TD-DETR~\cite{zhou2025devil} (ICCV'2025) & - & - & - & - & - & \underline{60.9} & \textbf{40.4} & - \\
\rowcolor{cyan!10}
MPAR (Ours) &  \textbf{56.0} & \textbf{41.5} & \textbf{24.8} & \textbf{38.8} & \textbf{73.1} & \textbf{61.2} & \underline{40.0} &  \textbf{52.6}  \\

    \bottomrule
  \end{tabular}
  }
  % \vspace{-0.5cm}
\end{table*}

\section{Experiments}
\subsection{Dataset}

\noindent\textbf{QVHighlights}~\cite{lei2021detecting} dataset contains 10,148 videos, 18,367 annotated moments, and 10,310 sentences, with each moment rated on a five-level highlight scale from \enquote{very poor} to \enquote{excellent}. It supports both video grounding and highlight detection, with fair evaluation ensured by withholding test-set annotations. We follow the standard protocol and report results on the official validation set.
% dataset contains 10,148 videos, 18,367 annotated moments, and 10,310 sentences, each moment labeled on a five-level highlight scale from “very poor” to “excellent.” It supports both video grounding and highlight detection, with fair evaluation ensured by withholding test-set annotations. In our experiments, we follow the standard protocol and report results on the official validation set.
% is a recently proposed benchmark that supports both video grounding and highlight detection by selecting representative moments within a video. It contains 10,148 videos, 18,367 annotated moments, and 10,310 sentences. Each moment is annotated with a five-level highlight rating (ranging from “very poor” to “excellent”) to facilitate highlight detection. As the ground-truth annotations for the test set are not publicly available, the evaluation process is particularly fair. In our experiments, we report results on the official validation set, following the standard protocol.
\newline
\textbf{Charades-STA}~\cite{gao2017tall} dataset comprises 9,848 indoor videos and 16,128 natural language queries, each aligned with a specific temporal segment. Videos average 30 seconds in length, with annotated moments around 8.1 seconds. The official split provides 12,408 training pairs and 3,720 testing pairs, allowing consistent evaluation of temporal moment retrieval models.
% dataset contains 9,848 indoor videos and 16,128 natural language queries, each paired with a specific temporal segment. Videos average 30 seconds, with annotated moments around 8.1 seconds long. The official split provides 12,408 moment-query pairs for training and 3,720 pairs for testing, supporting consistent evaluation of temporal moment retrieval models.
% is a widely used benchmark for temporal moment retrieval tasks. It consists of 9,848 videos capturing everyday indoor activities and 16,128 human-annotated natural language queries. Each query is aligned with a specific temporal segment (moment) within the corresponding video, allowing models to learn fine-grained associations between language and time-localized visual content. On average, the full videos have a duration of 30 seconds, while the annotated moments last around 8.1 seconds, offering a diverse temporal range for training. The dataset provides an official split with 12,408 moment-query pairs for training and 3,720 pairs for testing, enabling consistent and reproducible evaluation across different approaches.
\newline
\textbf{TACoS}~\cite{regneri2013grounding} dataset consists of high-resolution cooking videos with 17,334 action descriptions, covering 146,771 words and 75,210 lexical instances. Video durations range from 1 to 23 minutes (average 4.5 minutes), offering diverse temporal spans for fine-grained video language alignment. The official split includes 10,146 training, 4,589 validation, and 4,083 testing pairs for standardized evaluation.

\subsection{Metrics and Implementation Details}
\noindent\textbf{Metrics.} We use standard evaluation metrics from TR-DETR~\cite{sun2024tr}, TD-DETR~\cite{zhou2025devil}, and QD-DETR~\cite{moon2023query}. For QVHighlights, we report Recall$@$1 at IoU 0.5 and 0.7, and mAP at IoU 0.5 and 0.75, as well as average mAP over 10 thresholds from 0.5 to 0.95~\cite{lei2021detecting}. For TACoS and Charades-STA, we evaluate Recall$@$1 with IoU $\in$ $\left\{0.3, 0.7\right\}$. 
% We adopt widely-used evaluation metrics from recent studies such as TR-DETR~\cite{sun2024tr}, Moment-DETR~\cite{lei2021detecting} and QD-DETR~\cite{moon2023query}. For QVHighlights, we report Recall$@$1 with IoU $\in$ $\left\{0.5, 0.7\right\}$ and mean average precision (mAP) with IoU $\in$  $\left\{0.5, 0.75\right\}$. Following ~\cite{lei2021detecting}, we also compute mAP by averaging over 10 evenly sampled IoU thresholds between 0.5 and 0.95, referred to as average mAP. For Charades-STA and TACos, we evaluate Recall$@$1 with IoU $\in$ $\left\{0.3, 0.7\right\}$. The hidden dimension $d\_modal$ is set to 256.
\newline
\textbf{Implementation Details.}
For QVHighlights, visual features are extracted from SlowFast~\cite{feichtenhofer2019slowfast} and CLIP~\cite{radford2021learning}, while textual features are obtained via CLIP’s text encoder~\cite{radford2021learning}. The model is trained for 200 epochs with a batch size of 32 and a learning rate of 1e-4.
For Charades-STA, visual features follow TR-DETR~\cite{sun2024tr} and are extracted using VGG~\cite{sengupta2019going}, I3D~\cite{carreira2017quo}, SlowFast~\cite{feichtenhofer2019slowfast}, and CLIP~\cite{radford2021learning}, with textual features from GloVe~\cite{pennington2014glove}. Training runs for 100 epochs with a batch size of 8 and a learning rate of 1e-4.
For TACoS, visual features are extracted using C3D~\cite{tran2015learning}, and textual features are derived from GloVe~\cite{pennington2014glove}. The model is trained for 200 epochs with a batch size of 32 and a learning rate of 2$\times$1e-4. The hidden dimension $d_{modal}$ is set to 256.
All experiments are conducted on a single NVIDIA RTX 3090 GPU.
% For the QVHighlights dataset, we utilize visual features extracted from SlowFast~\cite{feichtenhofer2019slowfast} and CLIP~\cite{radford2021learning}, while textual features are obtained using the text encoder from CLIP~\cite{radford2021learning}. The model is trained for 200 epochs with a batch size of 32 and a learning rate of 1e-4.
% In the case of Charades-STA, we extract visual features with VGG~\cite{sengupta2019going}, I3D~\cite{carreira2017quo}, SlowFast~\cite{feichtenhofer2019slowfast}, and CLIP~\cite{radford2021learning} following TR-DETR~\cite{sun2024tr}. Textual features are derived from GloVe~\cite{pennington2014glove}. The training process spans 100 epochs, with a batch size of 8 and a learning rate of 1e-4.
% In the case of TACos, we extract visual features using C3D~\cite{tran2015learning}. Textual features are derived from GloVe~\cite{pennington2014glove}. The training process spans 200 epochs, with a batch size of 32 and a learning rate of 2$\times$1e-4. All experiments are conducted on a single NVIDIA RTX 3090 GPU.

\subsection{Comparison with State-of-the-art Methods}
% Table~\ref{QVHighlights val} presents the moment retrieval performance of MPAR on the QVHighlights dataset. Compared with several state-of-the-art methods, MPAR achieves consistently superior Recall scores. Although the mAP performance is slightly lower than TD-DETR, our method demonstrates clear advantages in highlight detection. These results indicate that incorporating multi-perspective auxiliary representation and cross-modal associations effectively enhances semantic–visual alignment, mitigates semantic drift, and improves temporal localization accuracy.

Table~\ref{QVHighlights val} reports the performance comparison on the QVHighlights dataset. MPAR achieves strong recall performance, obtaining the best R1@0.5 and R1@0.7 scores among the compared methods. This indicates that the proposed multi-perspective auxiliary representations effectively improve the model's ability to retrieve text-relevant temporal segments, especially when the target moment contains diverse visual cues that are not fully specified by the original query. Meanwhile, MPAR achieves competitive mAP results. We observe that its Avg. mAP is slightly lower than the best-performing method, which may be attributed to the fact that MPAR emphasizes semantic coverage and target relevance. While such a design helps retrieve more semantically related moments, further boundary refinement may still be required to improve high-IoU localization precision. Overall, the results show that MPAR effectively enhances semantic-visual alignment and alleviates semantic drift caused by sparse textual descriptions.

Furthermore, Table~\ref{TACoS and Charades-STA} reports results on TACoS and Charades-STA. MPAR achieves competitive or superior performance across both datasets, demonstrating that the proposed auxiliary representation learning strategy generalizes well to different video domains, query styles, and temporal moment distributions.

% However, although our improvements on QVHighlights are substantial, the gains on TACoS and Charades-STA are relatively limited. We attribute this to the limited temporal complexity and less diverse query-video interactions in these datasets, degrading the model’s ability to learn generalized anchor-query correspondences, as also noted by RGTR~\cite{sun2025diversifying}.

% \usepackage{subcaption} % 请确保导言区已包含此包

\subsection{Ablation Studies}

\noindent\textbf{Effect of each component.}
Table~\ref{tab:ablations_MPSA} summarizes the ablation results of VSAU, SDG, and MPSA on QVHighlights. The first four rows show variants without MPSA, while the last four rows include MPSA, enabling us to examine both individual component effects and their interactions.

The results show that using VSAU or SDG alone does not consistently improve all metrics, indicating that global subject consistency and local visual diversification are complementary rather than independently sufficient. When VSAU and SDG are jointly used, the Avg. mAP improves from 45.79 to 46.25 without MPSA, demonstrating that VISS enriches auxiliary representations with more informative visual context. With MPSA enabled, the full model achieves the best Avg. mAP of 47.15, showing that MPSA further converts visually grounded auxiliary representations into effective temporal alignment signals. Overall, these results verify that VISS and MPSA play complementary roles in improving cross-modal alignment and moment localization.

\begin{table}[h]
\centering
\caption{Effectiveness of VISS and MPSA modules on QVHighlights.}
\label{tab:ablations_MPSA}

\setlength{\tabcolsep}{5pt}
\renewcommand{\arraystretch}{1.05}

\begin{tabular}{c|c|c|ccc}
\toprule
Methods & VSAU & SDG & R1@0.5 & R1@0.7 & Avg. mAP \\
\midrule

\multirow{4}{*}{w/o MPSA}
& - & - & 67.48 & 51.23 & 45.79 \\
& - & \checkmark & 68.26 & 51.35 & 45.07 \\
& \checkmark & - & 68.71 & 51.68 & 45.28 \\
& \checkmark & \checkmark & 69.16 & 53.29 & 46.25 \\
\midrule

\multirow{4}{*}{w/ MPSA}
& - & - & 67.29 & 52.00 & 45.78 \\
& - & \checkmark & 68.65 & 52.00 & 45.49 \\
& \checkmark & - & 68.19 & 52.26 & 45.91 \\
& \checkmark & \checkmark & 69.61 & 53.68 & 47.15 \\
\bottomrule
\end{tabular}

\end{table}

\noindent
\textbf{Effect of AR number and text length $k^i$.} Table~\ref{tab:ablations_AR} shows that dual ARs with token lengths \{4, 10\} achieve optimal performance. This confirms that a well-configured auxiliary set effectively captures diverse semantic cues, thereby enhancing temporal localization.
\begin{table}[h]
\centering
\caption{Impact of Auxiliary Representation (AR) settings on QVHighlights.}
\label{tab:ablations_AR}

\setlength{\tabcolsep}{5pt}
\renewcommand{\arraystretch}{1.05}

\begin{tabular}{c|c|ccc}
\toprule
$N_a$ & Tokens $k^i$ & R1@0.5 & R1@0.7 & Avg. mAP \\
\midrule

w/o AR & - & 67.48 & 51.23 & 45.79 \\
\midrule

\multirow{2}{*}{1}
& \{4\} & 67.29 & 52.52 & 45.23 \\
& \{10\} & 68.00 & 53.10 & 46.47 \\
\midrule

\multirow{3}{*}{2}
& \{4,4\} & 68.00 & 51.35 & 45.68 \\
&  \{4,10\} & 69.61 & 53.68 & 47.15 \\
& \{10,14\} & 67.10 & 51.81 & 45.63 \\
\midrule

\multirow{2}{*}{3}
& \{4,4,4\} & 68.45 & 51.94 & 45.87 \\
& \{4,10,14\} & 68.13 & 53.55 & 46.48 \\
\bottomrule
\end{tabular}

\end{table}

\begin{figure*}[t]
  \centering
   \includegraphics[width=0.9\linewidth]{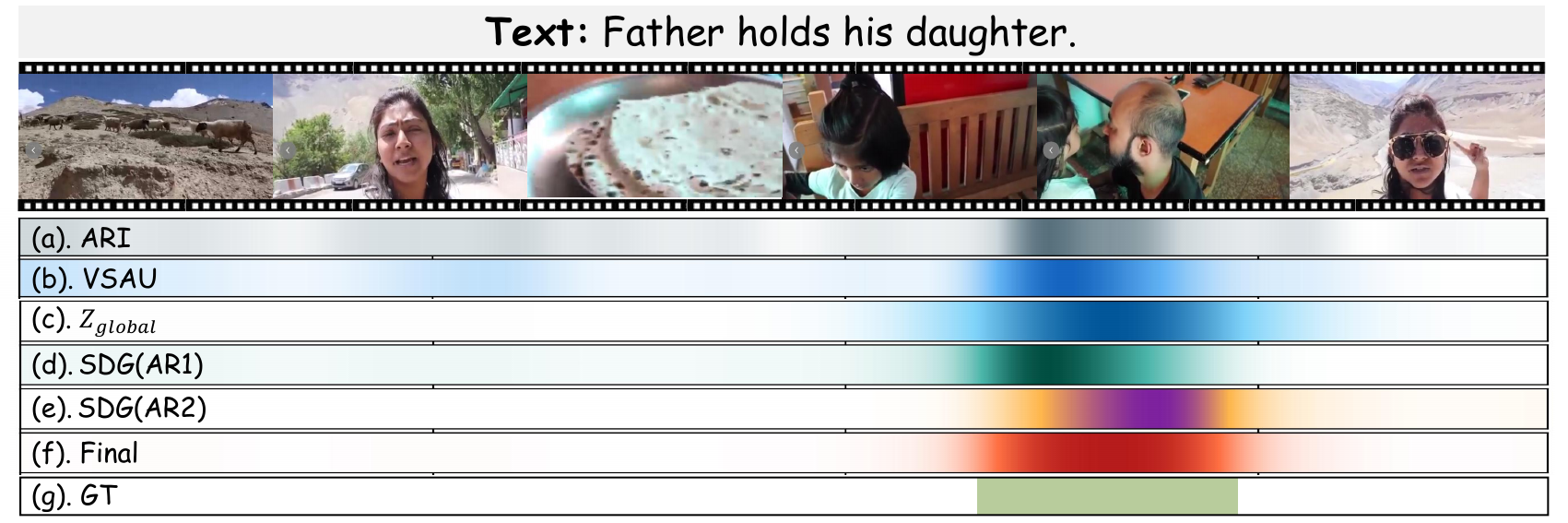}
   \caption{Visualization of module-level attention maps.}
   \label{fig:module analy}
   % \vspace{-0.6cm}
\end{figure*}

\noindent
\textbf{Semantic validity.}
Table~\ref{tab:sim_analysis} analyzes the semantic relevance and diversity of the learned auxiliary representations. Specifically, we compute the cosine similarity between each auxiliary representation and the global semantic anchor $z_{global}$, as well as the similarity between different auxiliary representations. The high similarities between $z_{global}$ and $AR_1$/$AR_2$ (0.89 and 0.94) indicate that the generated auxiliary representations remain semantically consistent with the original query, avoiding severe semantic drift. Meanwhile, the inter-AR similarity of 0.77 suggests that different auxiliary representations are not identical, but retain moderate variation while sharing the same target semantics. These results show that MPAR can maintain semantic coherence while introducing controlled representation diversity, which provides a reliable basis for subsequent visual cue injection and cross-modal alignment.
\begin{table}[h]
\centering
\caption{Semantic relevance and diversity of auxiliary representations.}
\label{tab:sim_analysis}

\setlength{\tabcolsep}{6pt}

\begin{tabular}{l c c}
\toprule
Metric & Pair & Similarity \\
\midrule
Relevance & $z_{global}$ vs. $AR_1$ & 0.89 \\
Relevance & $z_{global}$ vs. $AR_2$ & 0.94 \\
\midrule
Diversity & $AR_1$ vs. $AR_2$ & 0.77 \\
\bottomrule
\end{tabular}

\end{table}
% \noindent
% \textbf{Semantic validity.}
% Table~\ref{tab:sim_analysis} provides quantitative evidence that MPAR’s  performance gains stem from semantic complementarity rather than simple regularization. To ensure coherence, $z_{global}$ acts as a semantic anchor, the high AR-$z_{global}$ consistency (0.89/0.94) ensures relevance, while moderate inter-AR similarity (0.77) confirms successful perspective disentanglement. 
% % Visualizations in Fig.~\ref{fig:module analy} further show that while $z_{global}$ maintains global context, individual ARs focus on distinct temporal facets. 

\noindent
\textbf{VISS impact verification.}
Table~\ref{tab:viss_verification} further verifies whether the performance gain comes from visually-induced semantic enrichment or merely from the regularization effect introduced by random masking. The variant \enquote{Mask w/o VISS} only applies the masking strategy in ARI, where auxiliary representations are generated from the original textual features without incorporating visual cues. In contrast, \enquote{Mask w/ VISS} further introduces visual-induced semantic structuring, allowing the auxiliary representations to absorb target-specific visual information from the video.

As shown in the table, adding VISS consistently improves all metrics. These improvements indicate that random masking alone is insufficient to produce effective auxiliary representations. Instead, the visual cues introduced by VISS play a crucial role in enriching the initially constructed representation slots with target-related semantics. This confirms that the gains of MPAR mainly stem from visually grounded semantic enhancement rather than masking-induced stability.

\begin{table}[h]
\centering
\caption{Effect of VISS-driven enrichment vs. random masking.}
\label{tab:viss_verification}

\setlength{\tabcolsep}{6pt}
\renewcommand{\arraystretch}{1.15}

\begin{tabular}{l c c c}
\toprule
Method & R1@0.5 & R1@0.7 & Avg. mAP \\
\midrule
Mask w/o VISS & 67.29 & 52.00 & 45.78 \\
\rowcolor{cyan!10}
Mask w/ VISS & 69.61 & 53.68 & 47.15 \\
\bottomrule
\end{tabular}

\end{table}

\noindent
\textbf{Effect of dropout mask $p_i$.} Table~\ref{tab:ablations_Dropout} reveals that a moderate dropout probability of \{0.2, 0.15\} in ARI yields optimal results. This indicates that balanced stochastic masking effectively enhances model generalization and temporal localization accuracy.

\begin{table}[h]
  \centering
   \caption{Ablation study of the dropout mask probability in Auxiliary Representation Initialization.}
  \label{tab:ablations_Dropout}
  \resizebox{0.8\linewidth}{!}{
\begin{tabular}{c|ccc}

\toprule
Dropout mask $p_i$ & R1$@$0.5 & R1$@$0.7 & Avg. mAP \\
\hline
w/o dropout & 68.13 & 51.74  & 45.19 \\
% \hdashline
\{0.4, 0.3\}& 68.00 & 52.06 & 46.77 \\
\rowcolor{cyan!10}
\{0.2, 0.15\}& 69.61 & 53.68 & 47.15 \\
\{0.1, 0.075\} & 67.68 & 52.13  & 44.97 \\
\bottomrule
\end{tabular}
}

\end{table}

\noindent
\textbf{Effect of threshold $\theta$.} Table~\ref{tab:ablations_threshold} reveals that using the mean similarity as the threshold $\theta$ yields optimal results. This strategy effectively balances the inclusion of target-relevant segments with the exclusion of noise, thereby enhancing representation quality.
\begin{table}[h]
\centering
 \caption{Ablation study of the Threshold $\theta$ for target-related segments in Semantic Diversification Generator~(SDG).}
  \label{tab:ablations_threshold}
\resizebox{0.75\linewidth}{!}{
\begin{tabular}{c|ccc}
\toprule
Threshold $\theta$ & R1$@$0.5 & R1$@$0.7 & Avg. mAP \\
\hline
w/o threshold & 68.13 & 52.84 & 46.27 \\
% \hdashline
top-10 & 67.81 & 50.84 & 44.90 \\
top-30 & 67.42 & 51.94 & 45.25 \\
\rowcolor{cyan!10}
mean value & 69.61 & 53.68 & 47.15 \\
\bottomrule
\end{tabular}
}

\end{table}

\noindent
\textbf{Effect of the Concept tokens $N_c$.}
Table~\ref{tab:ablations_Concept tokens} indicates that 64 concept tokens $N_c$ yield optimal performance, striking a balance between capturing diverse visual semantics and preventing redundancy.

\begin{table}[h]
  \centering
   \caption{Ablation study of the Concept tokens $N_c$ in Semantic Diversification Generator~(SDG).}
  \label{tab:ablations_Concept tokens}
\resizebox{0.85\linewidth}{!}{
\begin{tabular}{c|ccc}
\toprule
Concept tokens $N_c$ & R1$@$0.5 & R1$@$0.7 & Avg. mAP \\
\hline
% \hdashline
32 & 67.61 &  51.10 & 45.04 \\
\rowcolor{cyan!10}
64 & 69.61 & 53.68 & 47.15 \\
128 & 66.13 & 51.42 & 46.21\\
\bottomrule
\end{tabular}
}

\end{table}

\noindent
\textbf{Inference efficiency.}
As shown in Table~\ref{tab:efficiency}, MPAR introduces only a slight increase in parameters and memory usage compared with the baseline, while maintaining nearly the same computational cost. The additional runtime mainly comes from processing multiple auxiliary representations and their cross-modal interactions. Overall, MPAR achieves a reasonable trade-off between efficiency and performance, improving moment localization with limited inference overhead.
\begin{table}[h]
    \centering
    % \small
     % \vspace{-0.5cm}
    \caption{Inference efficiency.}
  \label{tab:efficiency}
\resizebox{\linewidth}{!}{%
\begin{tabular}{lcccc}
\toprule
 & Params(M) & FLOPs(G)& Runtime(ms) & Memory(MB) \\ \hline
Baseline       &     10.02         &  0.73          &    26.56           &  639.65              \\
MPAR  &  10.08            &     0.73       &     47.36          &   648.27  \\ \bottomrule
\end{tabular}
}
 % \vspace{-1cm}
\end{table}

\noindent\textbf{Analysis of supervision and constraint mechanisms.}
We emphasize that token learning in our framework does not start from random optimization, but is built upon pretrained semantic representations. Except for the randomly initialized $z_{global}$, all inputs are derived from pretrained text and video encoders, which already encode strong semantic priors. Therefore, token learning mainly focuses on semantic reorganization and cross-modal alignment rather than low-level feature discovery. For Concept Tokens, they are generated from raw video features and naturally inherit visual grounding cues. As shown in the figure, even without explicit supervision, Concept Tokens tend to attend to key action regions, indicating emergent semantic specialization. \begin{table}[h] 
\centering \small % \vspace{-0.5cm}
\caption{Analysis of Supervision and Constraint Mechanisms.} \label{tab:loss supervise} \begin{tabular}{l|ccc} \toprule & R1@0.5 & R1@0.7 & mAP \\ \hline $L_{orth}$ & 68.19 & 52.19 & 45.62 \\ $L_{cons}$ & 65.68 & 48.45 & 42.55 \\ \hline Ours & 69.61 & 53.68 & 47.15 \\ \bottomrule \end{tabular} % \vspace{-1cm}
\end{table}
To further examine whether explicit constraints are necessary, we additionally evaluate two auxiliary regularization terms, i.e., an orthogonality loss $L_{orth}$ and a consistency loss $L_{cons}$ in Table~\ref{tab:loss supervise}. Given the learned concept tokens $\mathbf{C}=[\mathbf{c}_1,\mathbf{c}_2,\ldots,\mathbf{c}_{N_c}]$, we first normalize each token and define the orthogonality loss as: 
\begin{equation} L_{orth} = \frac{1}{N_c(N_c-1)} \sum_{i=1}^{N_c}\sum_{j=1,j\neq i}^{N_c} \left(\frac{\mathbf{c}_i^\top \mathbf{c}_j}{\|\mathbf{c}_i\|_2\|\mathbf{c}_j\|_2}\right)^2. \end{equation} 
This loss penalizes high pairwise similarity among concept tokens, encouraging them to capture diverse visual semantics. For semantic consistency, we constrain each auxiliary representation to preserve the global semantics of the original query. Let $\bar{\mathbf{A}}^i$ denote the pooled representation of the $i$-th auxiliary representation, and $\mathbf{z}_t$ denote the global text token extracted from the original query. The consistency loss is formulated as: 
\begin{equation} L_{cons} = \frac{1}{N_a}\sum_{i=1}^{N_a} \left(1 - \frac{(\bar{\mathbf{A}}^i)^\top \mathbf{z}_t} {\|\bar{\mathbf{A}}^i\|_2, \|\mathbf{z}_t\|_2} \right).
\end{equation} 
This loss encourages all auxiliary representations to remain semantically aligned with the original query, thereby reducing semantic drift.

We report the results of these two constraints in Table~\ref{tab:loss supervise}. Although $L_{orth}$ encourages token diversity and $L_{cons}$ promotes semantic consistency, the performance does not surpass our default design. This indicates that MPAR can naturally learn diverse yet coherent auxiliary representations through pretrained semantic priors and visual-induced structure, without relying on additional explicit constraint terms.

\noindent\textbf{Analysis of Modal Encoder Design.}
Table~\ref{tab:Implementation of the Modal Encoder} compares different Modal Encoder designs. MLP- and CNN-based encoders show limited ability to model cross-modal interactions, leading to inferior localization performance. In contrast, the MHA-based encoder adaptively aggregates informative text-video features and achieves the best results, demonstrating its effectiveness for cross-modal fusion in moment retrieval.

\begin{table}[h]
\centering
\caption{Implementation of the Modal Encoder.}
% \caption{Comparison of different encoder designs in the SDG module on QVHighlights dataset.}
\label{tab:Implementation of the Modal Encoder}

\small
\begin{tabular}{l|ccc}
\toprule
 & {R1@0.5} & {R1@0.7} & {Avg. mAP} \\ \hline
MLP-based      &   66.00         &       52.45    &46.88\ \\
CNN-based        &   67.23    & 51.94  &       46.45       \\ \hline
{MHA-based (Ours)}     & {69.61}  & {53.68}  & {47.15}    \\ \bottomrule
\end{tabular}
\end{table}

\noindent\textbf{Analysis across different temporal durations.}
We analyze model performance across different temporal durations by dividing samples into short ($\leq$10s), medium (10--30s), and long ($>$30s) intervals, as shown in Table~\ref{tab:duration_breakdown}.
The results indicate that MPAR yields more noticeable improvements on medium and long videos, where temporal semantics are more distributed and require richer contextual understanding. In these cases, the proposed auxiliary representations help mitigate semantic ambiguity by incorporating complementary visual cues, leading to more accurate moment localization.
In contrast, performance gains on short videos are relatively limited, as these cases typically involve compact temporal boundaries and less contextual complexity, leaving less room for multi-perspective representation benefits.

\begin{table}[h]
\caption{Performance analysis across different temporal durations. We categorize samples into short ($\leq$10s), medium (10--30s), and long ($>$30s) intervals.}
\centering
\label{tab:duration_breakdown}
\small
\begin{tabular}{l|c|cc}
\hline
 & Ratio & Baseline & Ours\\ \hline
Short Samples (Hard)     & $\sim$21.9\%      & 11.07    &11.18   \\
Medium  Samples          & $\sim$48.8\%     & 50.28        & 51.57     \\
Long Samples          & $\sim$29.3\%     & 52.80        & 53.77     \\\hline
\end{tabular}
\end{table}

\noindent\textbf{Generalization analysis of the number of ARs.}
We further analyze the effect of the number of auxiliary representations (AR) in Table~\ref{tab:ar_generalization_transposed}. The results remain stable across different AR settings on both TACoS and Charades-STA, with only minor fluctuations as the number of representations changes. This indicates that MPAR is robust to the configuration of auxiliary views, and its gains mainly come from effective multi-perspective representation modeling rather than sensitive hyperparameter tuning.

\begin{table}[h]
\caption{Generalization analysis with varying numbers of auxiliary representations (AR). Results on TACoS and Charades-STA show the robustness of MPAR to different AR configurations.}
\centering
\label{tab:ar_generalization_transposed}
\small
\begin{tabular}{l|cccc}
\hline
 &   AR=1 & AR=2 & AR=3 & AR=4 \\ \hline
TACoS
 &38.55   & 38.80  & 38.57 & 38.99 \\ 
Charades-STA
 & 52.10  &  52.60 & 51.05 & 52.21 \\ \hline
\end{tabular}
\end{table}

\subsection{Visualization and Qualitative Result}
% \noindent
\textbf{Module-level analysis.} 
To investigate each component's contribution to localization, we visualize the evolution of attention maps in Fig.~\ref{fig:module analy}: (a) w/ ARI only, (b) VSAU, (c) $z_{global}$ in SDG, (d) $AR_1$ in SDG, (e) $AR_2$ in SDG, (f) Final Attention and (g) GT. 
We clarify that ARI is not intended to introduce new information but to create \enquote{semantic vacancies} (placeholders).
With ARI alone, attention is diverse but fragmented. 
VSAU utilizes video instance features to anchor multiple auxiliary representations onto the event entities. In the SDG module, distinct ARs localize from decoupled perspectives, attending to diverse temporal cues. Meanwhile, $z_{global}$ acts as a global semantic anchor, constraining the ARs to ensure precise alignment through the synergy of local facets and global context. Ultimately, MPAR effectively mitigates single-word overfitting and addresses the inherent semantic sparsity of the original query.
% \noindent
% \textbf{Verify score in MPSA.} 
% FigureXXX shows the text-to-video verification scores in the Semantic-Guided Visual Verifier (MPSA) module.
\begin{figure*}[t]
  \centering
  \includegraphics[width=0.9\textwidth]{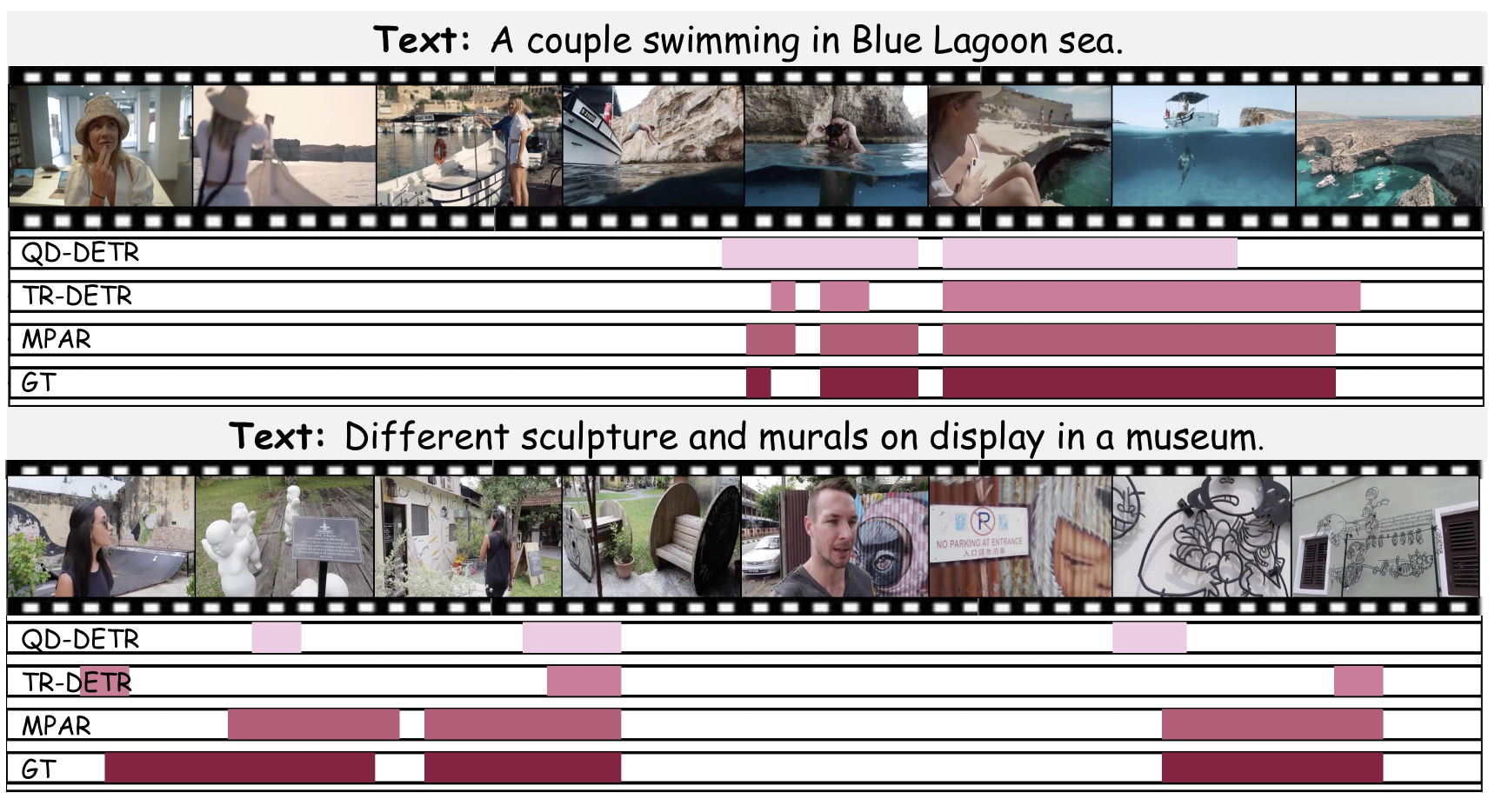}
  \caption{Qualitative results of MPAR on QVHighlights val split.}
  \label{keshihua}
  % \vspace{-0.7cm}
\end{figure*}

\begin{figure}[h]
    \centering
   
        \includegraphics[width=\linewidth]{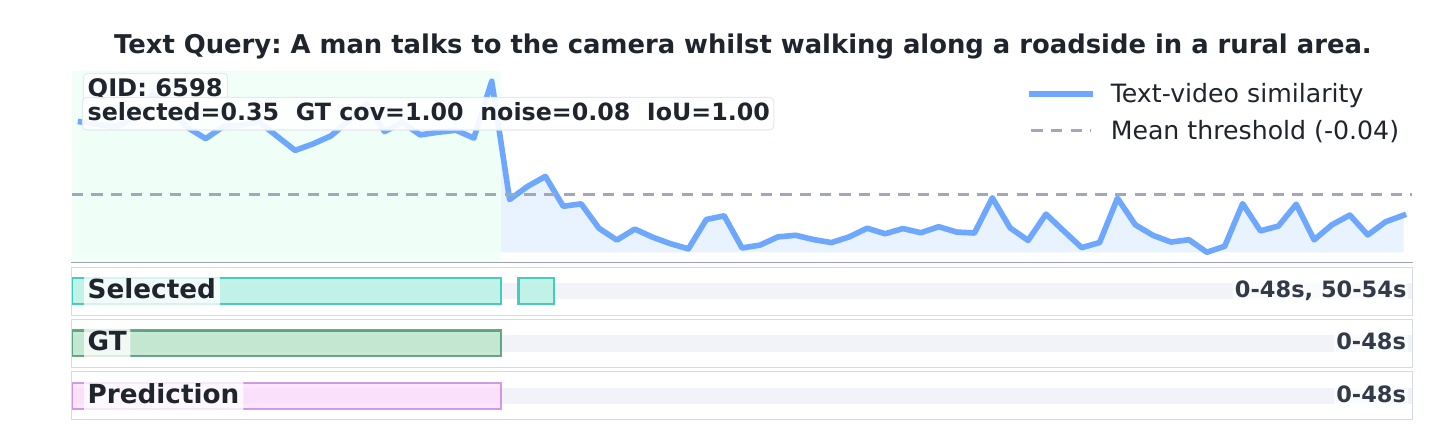}
        \label{fig:target_selection_case1}
    \vspace{-5mm}
        \includegraphics[width=\linewidth]{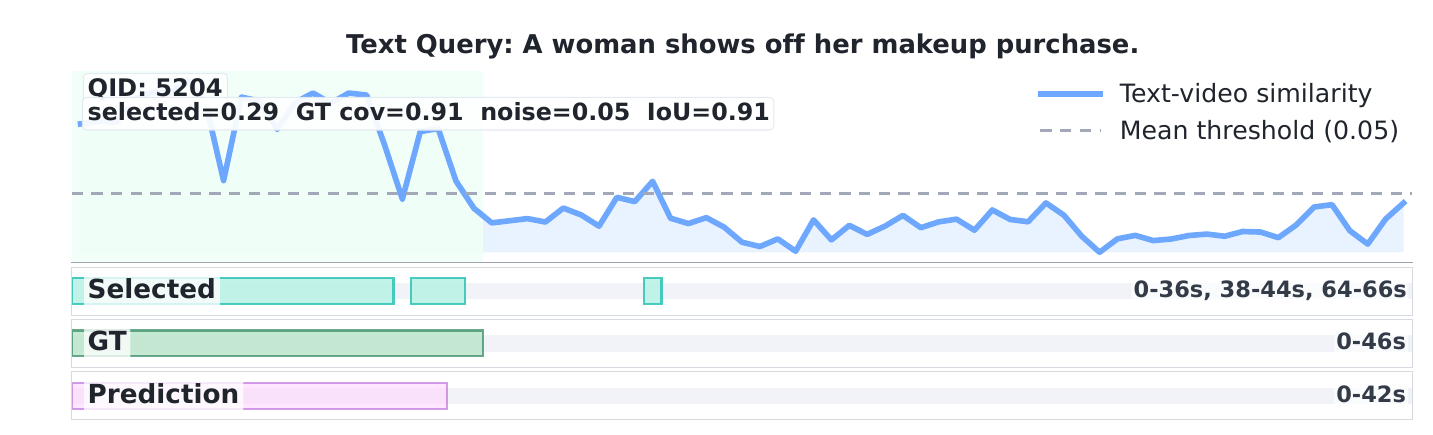}
        \label{fig:target_selection_case2}
        \vspace{-3mm}
\caption{Visualization of target-related segment selection in SDG.}
    \label{fig:target_segment_selection}
\end{figure}

\noindent\textbf{Target-related segment selection.}
Fig.~\ref{fig:target_segment_selection} visualizes the segment selection process in SDG. The selected segments largely overlap with the ground-truth moments in both cases, while introducing only limited noise. This indicates that SDG can effectively mine target-related visual cues to enrich auxiliary representations.

\begin{figure}[h]
\centering
\includegraphics[width=0.9\linewidth]{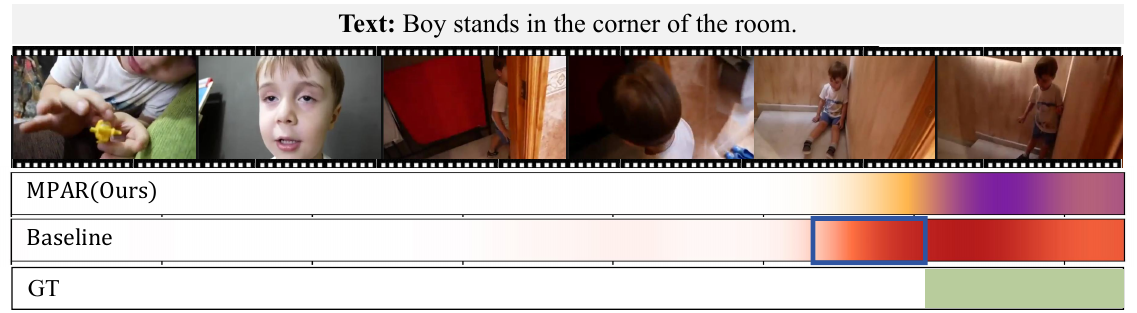}
\caption{Qualitative evidence of \enquote{bias/overfitting} motivation.}
\label{fig:bias_case}
% \vspace{-0.2cm}
\end{figure}

\noindent\textbf{Qualitative analysis of semantic bias.}
Fig.~\ref{fig:bias_case} shows a typical case of lexical bias in moment retrieval. For the query \enquote{Boy stands in the corner of the room}, the baseline over-relies on the noun \enquote{boy} and retrieves semantically mismatched segments. In contrast, MPAR better captures the predicate and spatial cues, such as \enquote{stands} and \enquote{corner}, by enriching auxiliary representations with visually induced cues. This suggests that MPAR can reduce word-level overfitting under sparse textual descriptions.

% \noindent
% \textbf{Attention maps of AR.} 
% Fig.~\ref{fig:attention} visualizes attention maps across different representations. The third row shows attention from the input text, while the fourth and fifth rows correspond to two auxiliary representations. The last row is the ground truth.
% Compared with the dispersed attention from the text alone, introducing auxiliary representations concentrates attention on the foreground and semantically relevant regions, better aligning with the ground truth. 
% % This demonstrates that auxiliary representations help the model focus on relevant visual content, improving temporal localization accuracy.

\noindent
\textbf{Qualitative Result.} Fig.~\ref{keshihua} presents qualitative results of MPAR on the QVHighlights dataset. We compare predictions from QD-DETR, TR-DETR, and MPAR. It can be observed that MPAR’s predicted temporal segments align more closely with the ground truth. This improvement can be attributed to the collaborative design and effective integration of the proposed modules, which together enhance the model’s temporal understanding and cross-modal alignment capabilities.

\section{Conclusion}
We propose a Multi-Perspective Auxiliary Representation framework for moment retrieval, aiming to enhance the expressiveness of a single textual query in complex video-language scenarios. Existing methods typically rely on one textual representation to guide localization, which may be insufficient to capture fine-grained visual attributes and can lead to ambiguous alignment when multiple semantically similar moments appear in the same video. To address this limitation, we first introduce Auxiliary Representation Initialization to construct multiple query-preserving auxiliary representation slots from the input text, providing flexible carriers for subsequent visual cue injection. We then design the Visual-Induced Semantic Structuring module to maintain a consistent target subject across these representations while incorporating target-specific visual cues, thereby improving the semantic completeness and visual grounding capability of the query representation. Finally, the Multi-Perspective Semantic Aligner integrates the enriched auxiliary representations with video features, guiding the model to focus on text-relevant temporal regions and reducing the semantic gap between sparse language descriptions and rich visual content. Extensive experiments on multiple moment retrieval benchmarks demonstrate the effectiveness of our framework.

%\section_{Proof of the First Zonklar Equation}
%\section_{Proof of the Second Zonklar Equation}

% \section{References Section}
% You can use a bibliography generated by BibTeX as a .bbl file.
%  BibTeX documentation can be easily obtained at:
%  http://mirror.ctan.org/biblio/bibtex/contrib/doc/
%  The IEEEtran BibTeX style support page is:
%  http://www.michaelshell.org/tex/ieeetran/bibtex/
 
% You can manually copy in the resultant .bbl file and set second argument of $\backslash${\tt{begin}} to the number of references
%  (used to reserve space for the reference number labels box).
\bibliographystyle{IEEEtran}
\bibliography{reference}

@String(AAAI = {AAAI})

@inproceedings{xu2019multilevel,
  title={Multilevel language and vision integration for text-to-clip retrieval},
  author={Xu, Huijuan and He, Kun and Plummer, Bryan A and Sigal, Leonid and Sclaroff, Stan and Saenko, Kate},
  booktitle={Proceedings of the AAAI Conference on Artificial Intelligence},
  volume={33},
  number={01},
  pages={9062--9069},
  year={2019}
}

@inproceedings{zhang2020learning,
  title={Learning 2d temporal adjacent networks for moment localization with natural language},
  author={Zhang, Songyang and Peng, Houwen and Fu, Jianlong and Luo, Jiebo},
  booktitle={Proceedings of the AAAI conference on artificial intelligence},
  volume={34},
  number={07},
  pages={12870--12877},
  year={2020}
}

@article{ghosh2019excl,
  title={Excl: Extractive clip localization using natural language descriptions},
  author={Ghosh, Soham and Agarwal, Anuva and Parekh, Zarana and Hauptmann, Alexander},
  journal={arXiv preprint arXiv:1904.02755},
  year={2019}
}

@article{zhang2021natural,
  title={Natural language video localization: A revisit in span-based question answering framework},
  author={Zhang, Hao and Sun, Aixin and Jing, Wei and Zhen, Liangli and Zhou, Joey Tianyi and Goh, Rick Siow Mong},
  journal={IEEE transactions on pattern analysis and machine intelligence},
  volume={44},
  number={8},
  pages={4252--4266},
  year={2021},
  publisher={IEEE}
}

@inproceedings{mun2020local,
  title={Local-global video-text interactions for temporal grounding},
  author={Mun, Jonghwan and Cho, Minsu and Han, Bohyung},
  booktitle={Proceedings of the IEEE/CVF Conference on Computer Vision and Pattern Recognition},
  pages={10810--10819},
  year={2020}
}

@inproceedings{carion2020end,
  title={End-to-end object detection with transformers},
  author={Carion, Nicolas and Massa, Francisco and Synnaeve, Gabriel and Usunier, Nicolas and Kirillov, Alexander and Zagoruyko, Sergey},
  booktitle={European conference on computer vision},
  pages={213--229},
  year={2020},
  organization={Springer}
}

@article{lei2021detecting,
  title={Detecting moments and highlights in videos via natural language queries},
  author={Lei, Jie and Berg, Tamara L and Bansal, Mohit},
  journal={Advances in Neural Information Processing Systems},
  volume={34},
  pages={11846--11858},
  year={2021}
}

@inproceedings{sun2024tr,
  title={Tr-detr: Task-reciprocal transformer for joint moment retrieval and highlight detection},
  author={Sun, Hao and Zhou, Mingyao and Chen, Wenjing and Xie, Wei},
  booktitle={Proceedings of the AAAI Conference on Artificial Intelligence},
  volume={38},
  number={5},
  pages={4998--5007},
  year={2024}
}

@inproceedings{gao2017tall,
  title={Tall: Temporal activity localization via language query},
  author={Gao, Jiyang and Sun, Chen and Yang, Zhenheng and Nevatia, Ram},
  booktitle={Proceedings of the IEEE international conference on computer vision},
  pages={5267--5275},
  year={2017}
}

@inproceedings{moon2023query,
  title={Query-dependent video representation for moment retrieval and highlight detection},
  author={Moon, WonJun and Hyun, Sangeek and Park, SangUk and Park, Dongchan and Heo, Jae-Pil},
  booktitle={Proceedings of the IEEE/CVF conference on computer vision and pattern recognition},
  pages={23023--23033},
  year={2023}
}

@inproceedings{feichtenhofer2019slowfast,
  title={Slowfast networks for video recognition},
  author={Feichtenhofer, Christoph and Fan, Haoqi and Malik, Jitendra and He, Kaiming},
  booktitle={Proceedings of the IEEE/CVF international conference on computer vision},
  pages={6202--6211},
  year={2019}
}

@inproceedings{carreira2017quo,
  title={Quo vadis, action recognition? a new model and the kinetics dataset},
  author={Carreira, Joao and Zisserman, Andrew},
  booktitle={proceedings of the IEEE Conference on Computer Vision and Pattern Recognition},
  pages={6299--6308},
  year={2017}
}

@inproceedings{pennington2014glove,
  title={Glove: Global vectors for word representation},
  author={Pennington, Jeffrey and Socher, Richard and Manning, Christopher D},
  booktitle={Proceedings of the 2014 conference on empirical methods in natural language processing (EMNLP)},
  pages={1532--1543},
  year={2014}
}

@inproceedings{radford2021learning,
  title={Learning transferable visual models from natural language supervision},
  author={Radford, Alec and Kim, Jong Wook and Hallacy, Chris and Ramesh, Aditya and Goh, Gabriel and Agarwal, Sandhini and Sastry, Girish and Askell, Amanda and Mishkin, Pamela and Clark, Jack and others},
  booktitle={International conference on machine learning},
  pages={8748--8763},
  year={2021},
  organization={PmLR}
}

@inproceedings{rezatofighi2019generalized,
  title={Generalized intersection over union: A metric and a loss for bounding box regression},
  author={Rezatofighi, Hamid and Tsoi, Nathan and Gwak, JunYoung and Sadeghian, Amir and Reid, Ian and Savarese, Silvio},
  booktitle={Proceedings of the IEEE/CVF conference on computer vision and pattern recognition},
  pages={658--666},
  year={2019}
}

@article{moon2023correlation,
  title={Correlation-Guided Query-Dependency Calibration for Video Temporal Grounding},
  author={Moon, WonJun and Hyun, Sangeek and Lee, SuBeen and Heo, Jae-Pil},
  journal={arXiv preprint arXiv:2311.08835},
  year={2023}
}

@inproceedings{jang2023knowing,
  title={Knowing where to focus: Event-aware transformer for video grounding},
  author={Jang, Jinhyun and Park, Jungin and Kim, Jin and Kwon, Hyeongjun and Sohn, Kwanghoon},
  booktitle={Proceedings of the IEEE/CVF International Conference on Computer Vision},
  pages={13846--13856},
  year={2023}
}

@article{regneri2013grounding,
  title={Grounding action descriptions in videos},
  author={Regneri, Michaela and Rohrbach, Marcus and Wetzel, Dominikus and Thater, Stefan and Schiele, Bernt and Pinkal, Manfred},
  journal={Transactions of the Association for Computational Linguistics},
  volume={1},
  pages={25--36},
  year={2013},
  publisher={MIT Press One Rogers Street, Cambridge, MA 02142-1209, USA journals-info~…}
}

@inproceedings{tran2015learning,
  title={Learning spatiotemporal features with 3d convolutional networks},
  author={Tran, Du and Bourdev, Lubomir and Fergus, Rob and Torresani, Lorenzo and Paluri, Manohar},
  booktitle={Proceedings of the IEEE international conference on computer vision},
  pages={4489--4497},
  year={2015}
}

@inproceedings{lin2023univtg,
  title={Univtg: Towards unified video-language temporal grounding},
  author={Lin, Kevin Qinghong and Zhang, Pengchuan and Chen, Joya and Pramanick, Shraman and Gao, Difei and Wang, Alex Jinpeng and Yan, Rui and Shou, Mike Zheng},
  booktitle={Proceedings of the IEEE/CVF International Conference on Computer Vision},
  pages={2794--2804},
  year={2023}
}

@inproceedings{yang2024task,
  title={Task-driven exploration: Decoupling and inter-task feedback for joint moment retrieval and highlight detection},
  author={Yang, Jin and Wei, Ping and Li, Huan and Ren, Ziyang},
  booktitle={Proceedings of the IEEE/CVF Conference on Computer Vision and Pattern Recognition},
  pages={18308--18318},
  year={2024}
}

@article{li2023momentdiff,
  title={Momentdiff: Generative video moment retrieval from random to real},
  author={Li, Pandeng and Xie, Chen-Wei and Xie, Hongtao and Zhao, Liming and Zhang, Lei and Zheng, Yun and Zhao, Deli and Zhang, Yongdong},
  journal={Advances in neural information processing systems},
  volume={36},
  pages={65948--65966},
  year={2023}
}

@article{tan2021augmented,
  title={Augmented 2d-tan: A two-stage approach for human-centric spatio-temporal video grounding},
  author={Tan, Chaolei and Lin, Zihang and Hu, Jian-Fang and Li, Xiang and Zheng, Wei-Shi},
  journal={arXiv preprint arXiv:2106.10634},
  year={2021}
}

@article{sengupta2019going,
  title={Going deeper in spiking neural networks: VGG and residual architectures},
  author={Sengupta, Abhronil and Ye, Yuting and Wang, Robert and Liu, Chiao and Roy, Kaushik},
  journal={Frontiers in neuroscience},
  volume={13},
  pages={95},
  year={2019},
  publisher={Frontiers Media SA}
}

@article{vaswani2017attention,
  title={Attention is all you need},
  author={Vaswani, Ashish and Shazeer, Noam and Parmar, Niki and Uszkoreit, Jakob and Jones, Llion and Gomez, Aidan N and Kaiser, {\L}ukasz and Polosukhin, Illia},
  journal={Advances in neural information processing systems},
  volume={30},
  year={2017}
}

@inproceedings{jiang2024prior,
  title={Prior knowledge integration via llm encoding and pseudo event regulation for video moment retrieval},
  author={Jiang, Yiyang and Zhang, Wengyu and Zhang, Xulu and Wei, Xiao-Yong and Chen, Chang Wen and Li, Qing},
  booktitle={Proceedings of the 32nd ACM International Conference on Multimedia},
  pages={7249--7258},
  year={2024}
}

@article{lu2019vilbert,
  title={Vilbert: Pretraining task-agnostic visiolinguistic representations for vision-and-language tasks},
  author={Lu, Jiasen and Batra, Dhruv and Parikh, Devi and Lee, Stefan},
  journal={Advances in neural information processing systems},
  volume={32},
  year={2019}
}

@article{edunov2018understanding,
  title={Understanding back-translation at scale},
  author={Edunov, Sergey and Ott, Myle and Auli, Michael and Grangier, David},
  journal={arXiv preprint arXiv:1808.09381},
  year={2018}
}

@article{wei2019eda,
  title={EDA: Easy data augmentation techniques for boosting performance on text classification tasks},
  author={Wei, Jason and Zou, Kai},
  journal={arXiv preprint arXiv:1901.11196},
  year={2019}
}

@article{gao2021simcse,
  title={Simcse: Simple contrastive learning of sentence embeddings},
  author={Gao, Tianyu and Yao, Xingcheng and Chen, Danqi},
  journal={arXiv preprint arXiv:2104.08821},
  year={2021}
}

@inproceedings{jindal2020augmenting,
  title={Augmenting NLP models using latent feature interpolations},
  author={Jindal, Amit and Chowdhury, Arijit Ghosh and Didolkar, Aniket and Jin, Di and Sawhney, Ramit and Shah, Rajiv},
  booktitle={Proceedings of the 28th International Conference on Computational Linguistics},
  pages={6931--6936},
  year={2020}
}

@inproceedings{zeng2019graph,
  title={Graph convolutional networks for temporal action localization},
  author={Zeng, Runhao and Huang, Wenbing and Tan, Mingkui and Rong, Yu and Zhao, Peilin and Huang, Junzhou and Gan, Chuang},
  booktitle={Proceedings of the IEEE/CVF international conference on computer vision},
  pages={7094--7103},
  year={2019}
}

@article{yang2020revisiting,
  title={Revisiting anchor mechanisms for temporal action localization},
  author={Yang, Le and Peng, Houwen and Zhang, Dingwen and Fu, Jianlong and Han, Junwei},
  journal={IEEE Transactions on Image Processing},
  volume={29},
  pages={8535--8548},
  year={2020},
  publisher={IEEE}
}

@inproceedings{liu2024stepwise,
  title={Stepwise Multi-grained Boundary Detector for Point-Supervised Temporal Action Localization},
  author={Liu, Mengnan and Wang, Le and Zhou, Sanping and Xia, Kun and Wu, Qi and Zhang, Qilin and Hua, Gang},
  booktitle={European Conference on Computer Vision},
  pages={333--349},
  year={2024},
  organization={Springer}
}

@article{zhou2025devil,
  title={The Devil is in the Spurious Correlations: Boosting Moment Retrieval with Dynamic Learning},
  author={Zhou, Xinyang and Wei, Fanyue and Duan, Lixin and Yao, Angela and Li, Wen},
  journal={arXiv preprint arXiv:2501.07305},
  year={2025}
}

@article{park2025language,
  title={Language-guided Recursive Spatiotemporal Graph Modeling for Video Summarization: J. Park et al.},
  author={Park, Jungin and Lee, Jiyoung and Sohn, Kwanghoon},
  journal={International Journal of Computer Vision},
  pages={1--25},
  year={2025},
  publisher={Springer}
}

@article{islam2025towards,
  title={Towards an Automated Multimodal Approach for Video Summarization: Building a Bridge Between Text, Audio and Facial Cue-Based Summarization},
  author={Islam, Md Moinul and Kakouros, Sofoklis and Heikkil{\"a}, Janne and Oussalah, Mourad},
  journal={arXiv preprint arXiv:2506.23714},
  year={2025}
}

@article{yan2024anomalysd,
  title={Anomalysd: Few-shot multi-class anomaly detection with stable diffusion model},
  author={Yan, Zhenyu and Fang, Qingqing and Lv, Wenxi and Su, Qinliang},
  journal={arXiv preprint arXiv:2408.01960},
  year={2024}
}

@article{enayati2024time,
  title={Time series anomaly detection via clustering-based representation},
  author={Enayati, Elham and Mortazavi, Reza and Basiri, Abdolali and Ghasemian, Javad and Moallem, Mahmoud},
  journal={Evolving Systems},
  volume={15},
  number={4},
  pages={1115--1136},
  year={2024},
  publisher={Springer}
}

@article{zhou2025optimization,
  title={Optimization of Multi dimensional Time Series Data Anomaly Detection Model Based on Graph Deviation Network and Convolutional Neural Network},
  author={Zhou, Yixin},
  journal={Procedia Computer Science},
  volume={261},
  pages={199--206},
  year={2025},
  publisher={Elsevier}
}

@article{he2024llms,
  title={Llms meet multimodal generation and editing: A survey},
  author={He, Yingqing and Liu, Zhaoyang and Chen, Jingye and Tian, Zeyue and Liu, Hongyu and Chi, Xiaowei and Liu, Runtao and Yuan, Ruibin and Xing, Yazhou and Wang, Wenhai and others},
  journal={arXiv preprint arXiv:2405.19334},
  year={2024}
}

@inproceedings{sandoval2025editduet,
  title={EditDuet: A Multi-Agent System for Video Non-Linear Editing},
  author={Sandoval-Castaneda, Marcelo and Russell, Bryan and Sivic, Josef and Shakhnarovich, Gregory and Caba Heilbron, Fabian},
  booktitle={Proceedings of the Special Interest Group on Computer Graphics and Interactive Techniques Conference Conference Papers},
  pages={1--11},
  year={2025}
}

@article{schmidt2025prior2former,
  title={Prior2Former--Evidential Modeling of Mask Transformers for Assumption-Free Open-World Panoptic Segmentation},
  author={Schmidt, Sebastian and K{\"o}rner, Julius and Fuchsgruber, Dominik and Gasperini, Stefano and Tombari, Federico and G{\"u}nnemann, Stephan},
  journal={arXiv preprint arXiv:2504.04841},
  year={2025}
}

@inproceedings{zhou2025hypidecoder,
  title={HyPiDecoder: Hybrid Pixel Decoder for Efficient Segmentation and Detection},
  author={Zhou, Fengzhe and Shi, Humphrey},
  booktitle={Proceedings of the IEEE/CVF International Conference on Computer Vision},
  pages={22100--22109},
  year={2025}
}

@inproceedings{sun2025diversifying,
  title={Diversifying query: Region-guided transformer for temporal sentence grounding},
  author={Sun, Xiaolong and Shi, Liushuai and Wang, Le and Zhou, Sanping and Xia, Kun and Wang, Yabing and Hua, Gang},
  booktitle={Proceedings of the AAAI Conference on Artificial Intelligence},
  volume={39},
  number={7},
  pages={7131--7139},
  year={2025}
}

@article{xie2025caption,
  title={Caption Assisted Multimodal Large Language Model for Video Moment Retrieval},
  author={Xie, Peiyu and Li, Jinxing and Lu, Guangming and Xu, Yong and Zhang, David},
  journal={IEEE Transactions on Image Processing},
  year={2025},
  publisher={IEEE}
}

@article{yang2022video,
  title={Video moment retrieval with cross-modal neural architecture search},
  author={Yang, Xun and Wang, Shanshan and Dong, Jian and Dong, Jianfeng and Wang, Meng and Chua, Tat-Seng},
  journal={IEEE Transactions on Image Processing},
  volume={31},
  pages={1204--1216},
  year={2022},
  publisher={IEEE}
}

@article{yang2021local,
  title={Local correspondence network for weakly supervised temporal sentence grounding},
  author={Yang, Wenfei and Zhang, Tianzhu and Zhang, Yongdong and Wu, Feng},
  journal={IEEE Transactions on Image Processing},
  volume={30},
  pages={3252--3262},
  year={2021},
  publisher={IEEE}
}

@article{sun2021maban,
  title={MABAN: Multi-agent boundary-aware network for natural language moment retrieval},
  author={Sun, Xiaoyang and Wang, Hanli and He, Bin},
  journal={IEEE Transactions on Image Processing},
  volume={30},
  pages={5589--5599},
  year={2021},
  publisher={IEEE}
}

@article{paul2021text,
  title={Text-based localization of moments in a video corpus},
  author={Paul, Sudipta and Mithun, Niluthpol Chowdhury and Roy-Chowdhury, Amit K},
  journal={IEEE Transactions on Image Processing},
  volume={30},
  pages={8886--8899},
  year={2021},
  publisher={IEEE}
}

@article{lin2020moment,
  title={Moment retrieval via cross-modal interaction networks with query reconstruction},
  author={Lin, Zhijie and Zhao, Zhou and Zhang, Zhu and Zhang, Zijian and Cai, Deng},
  journal={IEEE Transactions on Image Processing},
  volume={29},
  pages={3750--3762},
  year={2020},
  publisher={IEEE}
}

\end{document}